\documentclass{amsart}
\usepackage[dvips]{graphicx,psfrag}
\graphicspath{ {.} }
\setkeys{Gin}{width=\linewidth,keepaspectratio=true}
\usepackage{color}
\usepackage{url}
\usepackage{float}
\usepackage{epsfig}
\usepackage{algorithm}
\usepackage{algorithmic}
\usepackage[algo2e]{algorithm2e}
\usepackage{longtable}
\usepackage{subfigure}

\usepackage{natbib}
\bibpunct{(}{)}{;}{a}{,}{;}
\usepackage[utf8]{inputenc}
\usepackage[T1]{fontenc}
\usepackage[english]{babel}

\begin{document}
\title[BRKGA -- A Review]
{Biased 
Random-Key Genetic Algorithms: A Review}

\author[M.A. Londe]{M.A. Londe}
\address[Mariana A. Londe]
{Department of Industrial Engineering, PUC-Rio,\\
Rua Marqu\^es de S\~ao Vicente, 225, G\'avea  22453-900 Rio de Janeiro, RJ,
Brazil.}
\email[M.A. Londe]{mlonde@aluno.puc-rio.br}

\author[L.S. Pessoa]{L.S. Pessoa}
\address[Luciana S. Pessoa]{Department of Industrial Engineering, PUC-Rio,\\
Rua Marqu\^es de S\~ao Vicente, 225, G\'avea  22453-900 Rio de Janeiro, RJ,
Brazil.}
\email[L.S. Pessoa]{lucianapessoa@puc-rio.br}

\author[C. E. Andrade]
{C. E. Andrade}
\address[Cartlos A. Andrade]
{AT\&T Labs Research,\\
200 South Laurel Avenue, Middletown, NJ 07748 USA}
\email[C.E. Andrade]{cea@research.att.com}

\author[M.G.C. Resende]{M.G.C. Resende}
\address[Mauricio G.C. Resende]{Industrial and Systems Engineering,,
Univeristy of Washington,
3900 E Stevens Way NE,
Seattle, WA 98195 USA.}
\email[M.G.C. Resende]{mgcr@uw.edu}

\begin{abstract}
This paper is a comprehensive literature review of Biased Random-Key
Genetic Algorithms (BRKGA). BRKGA is a metaheuristic that employs
random-key-based chromosomes with biased, uniform, and elitist
mating strategies in a genetic algorithm framework. The review
encompasses over~$150$ papers with a wide range of applications,
including classical combinatorial optimization problems, real-world
industrial use cases, and non-orthodox applications such as neural
network hyperparameter tuning in machine learning. Scheduling is
by far the most prevalent application area in this review, followed
by network design and location problems. The most frequent hybridization
method employed is local search, and new features aim to increase
population diversity. Overall, this survey provides a comprehensive
overview of the BRKGA metaheuristic and its applications and
highlights important areas for future research.
\end{abstract}

\date{December 1, 2023.}
\thanks{Technical Report.}
%% cookie = ???????

\maketitle
\section{Introduction}
\label{Section:Introduction}

Genetic algorithms, first introduced by \citet{Holland1975:genetic_algorithm}, have proven to be effective for solving challenging optimization problems, ranging from discrete and combinatorial problems to non-linear and derivative-free optimization problems. This class of methods is a population-based metaheuristic that uses the principles of natural selection and survival of the fittest individuals, providing a powerful tool for searching large solution spaces. As a result, genetic algorithms have been applied in various domains due to their efficiency and versatility.

Biased Random-Key Genetic Algorithms (BRKGA) have become a popular variant of
genetic algorithms due to their successful application in several optimization problems, from classical hard combinatorial optimization 
problems \citep{Resende2012:steiner_triple_covering,Rochman2017:BRKGA_GD_CVRPTW}
to real-world problems, such as
packing \citep{Goncalves2011:multi_pop_constrained_2d_orthogonal_packing},
combinatorial auctions \citep{Andrade2015:winner_auctions},
scheduling \citep{%
Andrade2017:Managing_fota,%
Andrade2019:scheduling_software_cars,%
Andrade2019:flowshop_scheduling,%
Pessoa2018:flowshop_scheduling},
vehicle routing \citep{%
Andrade2013:Evolutionary_Algorithm_kIMDMTSP,%
Lopes2016:hub_location_routing},
clustering \citep{Andrade2014:Evol_Alg_Overlapping_Correlation_Clustering},
complex network design \citep{%
Andrade2022:PCI_Assignment_practical_opt,%
Andrade2015:wireless_backhaul},
placement of virtual machines in data
centers \citep{Stefanello2019:placement_virtual_machines},
and machine learning \citep{%
Caserta2016:data_fine_tuning,%
Paliwal2020:Reinfored_BRKGA_Comp_Graphs}, to cite only a few.
BRKGA was formally introduced by \citet{Goncalves2011:BRKGA}, although its
elements have appeared before in
\citep{%
Beirao1997:FirstBRKGA,%
Buriol2005:weight_setting_problem_OSPF_routing,%
Ericsson2002:Genetic_alg_OSPF,%
Goncalves1999:BRKGAFirst,%
Goncalves2002:hybrid_assembly_line_balancing%
}.

The first prominent feature of BRKGA is its agnosticism in solving the problem.
In most (meta) heuristic algorithms, the optimization mechanism is intrinsically tied to
the problem structure. Thus, while we apply the general framework, we still
need to develop or code details of the framework. In a BRKGA, such constant
rework is avoided using a standard representation of the solutions. The population in a BRKGA
lives in a half-open unit hypercube of dimension~$n$, and each solution or
individual is represented by a point in~$(0,1]^n$, called a \emph{chromosome}.
Such representation, proposed in \citep{Bean1994:random_keys}, makes the method
independent of the problem it solves and, therefore, allows for code reuse.
Such a strategy resembles 
modern machine learning algorithms: the knowledge representation is built over
a normalized matrix, and the so-called kernel functions are applied to measure
distances between points in that normalized
space \citep{Hofmann2008:Kernel_methods}. In a BRKGA, instead of having a kernel,
we have a \emph{decoder function} $f:[0,1)^n \to \mathcal{S}$ that maps
individuals from the BRKGA space to the problem solution space~$\mathcal{S}$.
Indeed, the decoder not only builds an actual solution from a chromosome but
also computes the solution value(s) used by the BRKGA as a measure of the quality
or fitness of the individual. We may see the decoder as the function that
computes the ``norm'' of a solution in the $[0,1)^n$ space. Such representation
allows a BRKGA to keep all evolutionary operators within the $[0,1)^n$ space, and
therefore, custom operators based on the problem structure are unnecessary.
This allows for fast prototyping and testing, thus reducing development costs.

The second outstanding feature of a BRKGA is its fast convergence to high-quality
solutions. Such achievement is due to the double elitism mechanism embedded in the evolutionary process of BRKGA. First, BRKGA 
hands over a subset of \emph{elite} individuals from one generation to the next, according to some performance metric (in general, the value(s) of
the objective function of the problem). The elite individuals are the best solutions for the current
generation. Excluding the very first iterations, a BRKGA will always have a set of
high-quality solutions in its population.  This
behavior contrasts with traditional genetic algorithms, which generally rebuild
the whole population every generation. Second, in the basic BRKGA, the mating
process occurs between a uniformly chosen individual from the elite set and a
uniformly chosen individual from the remaining population.  The combination of
such individuals is biased towards the elite individual, using uniform
crossover with probability~$\rho > 0.5$. Thus, there is a greater chance
of retaining substructures of a good solution while still allowing the insertion of substructures
of a not-so-good solution. The backdrop of double elitism is the fast
convergence to a local optimum. BRKGA deals with such issues by introducing
mutants (random solutions) and with other operators. Nevertheless, BRKGA
can deliver high-quality solutions in time-critical applications in a
reasonably short time, making it suitable for several industrial applications.

Due to such features, there has been an increasing number of applications of BRKGA in the
last years. 
In this paper, we present a detailed survey of the existing 
literature regarding BRKGA to discern the most studied problems, principal 
modifications to the framework, and most frequent hybridization methods 
explored. Over 150 academic articles were reviewed for this survey. 
We expect readers to have a concise but broad view of BRKGA applications
helping them in future endeavors.

% article sections

The remainder of this article is structured as follows. Section~\ref{Section:Tutorial} presents the fundamentals of the BRKGA framework in a detailed manner.
In 
Section~\ref{Section:Applications}, we present a comprehensive description of 
BRKGA applications. Section~\ref{Section:Hybrids} highlights the main 
hybridizations done with the framework. In Section~\ref{Section:Features}, we 
detail features that have been developed since the inception of this algorithm. 
Section~\ref{Section:Underperforming} comments about under-performing issues of 
BRKGA and their possible causes.
Possible directions for future research are presented in Section~\ref{Section:Agenda}.
Finally, in Section~\ref{Section:Conclusion}, we make concluding remarks.

\section{Fundamentals of the BRKGA}
\label{Section:Tutorial}

This section presents the basics of Biased Random Key Genetic Algorithms, as described in \citet{Goncalves2011:BRKGA}. 
Biased Random Key Genetic Algorithms (BRKGA) fall under the category of genetic algorithms, which are a class of search and optimization algorithms inspired by the process of natural selection. Introduced in the 1970s by  \citet{Holland1975:genetic_algorithm}, the fundamental concept of Genetic Algorithms (GA) involves treating a problem's solution as an individual within a population. These algorithms operate with a population of potential solutions, where each solution is represented as a chromosome or a string of genes. The population itself comprises a collection of these individuals.

GAs aim to replicate the process of evolution found in nature. Over time, the population evolves by implementing the Darwinian principle of survival of the fittest. Weaker individuals are unable to pass on their characteristics and gradually diminish within the population, while stronger individuals reproduce and transmit their characteristics. The algorithm proceeds through a series of generations, and at the conclusion of these generations, the individual with the best fitness value, often representing the best solution, is the output of the algorithm.

Individuals from one generation are combined to produce offspring that make up the next generation. The concept of passing on genetic information is generally implemented by selecting individuals to form pairs based on a probability proportional to their quality. This means that stronger individuals are more likely to be chosen for reproduction. The process of combining two solutions, the crossover, is the mechanism that intensifies the search, focusing on propagating the best traits of parents to their offspring. Once the two parents are selected, they are combined in some manner to generate offspring. An external process, akin to the mutations that occur in nature, introduces changes to the genetic information of the offspring. When a mutation is beneficial, it can lead to an improvement in the quality of the offspring.

In any search process, two essential components are at play: intensification and diversification. The crossover process contributes primarily to intensification, enhancing the exploration of promising areas in the solution space. On the other hand, diversification, responsible for exploring new regions of the solution space, is chiefly influenced by the process of mutation. The delicate balance between these phases is pivotal to the success of genetic algorithms in effectively optimizing complex problems.

This combination of selection, crossover, and mutation mimics the process of natural selection and drives the evolution of the population toward optimal or near-optimal solutions for the given problem.

\subsection{Genetic algorithms and random-keys}
\label{ss.tutorial.rk}

A random key is a randomly generated real number in the continuous interval $[0,1)$. Specifically, a vector $X$ composed of random--keys, often referred to simply as random keys, consists of an array containing $n$ such random keys.

Optimization problem solutions can be effectively encoded using these vectors of random keys, as proposed by \citet{Bean1994:random_keys}. Once a solution is encoded, a decoding process becomes necessary. The decoder is a deterministic algorithm that takes, as input, a vector of random keys and outputs a solution to the optimization problem and its cost. 

An example of a simple decoder applied to the Traveling Salesman Problem (TSP) follows. In this problem, there is a set of $n$ cities that must be visited exactly once, starting and ending in the same city. A solution to this problem is essentially a sequence of cities. Assume a vector of random keys denoted as $X = [0.234, 0.876, 0.321, 0.693, 0.087]$ is given. Each position in the vector corresponds to an index representing a city. To extract a solution for the TSP from this random key vector, one can simply sort this vector in increasing order and retrieve the sequence of indices associated with the sorted sequence. In this example, the sorted sequence yields the solution: 5-1-3-4-2, indicating the order in which the cities are visited by the traveling salesman. By applying the objective function to the decoded sequence, the decoder can determine the overall cost or fitness of the solution.

\begin{figure}[htb]
    \centering
    \includegraphics[scale = 0.4]{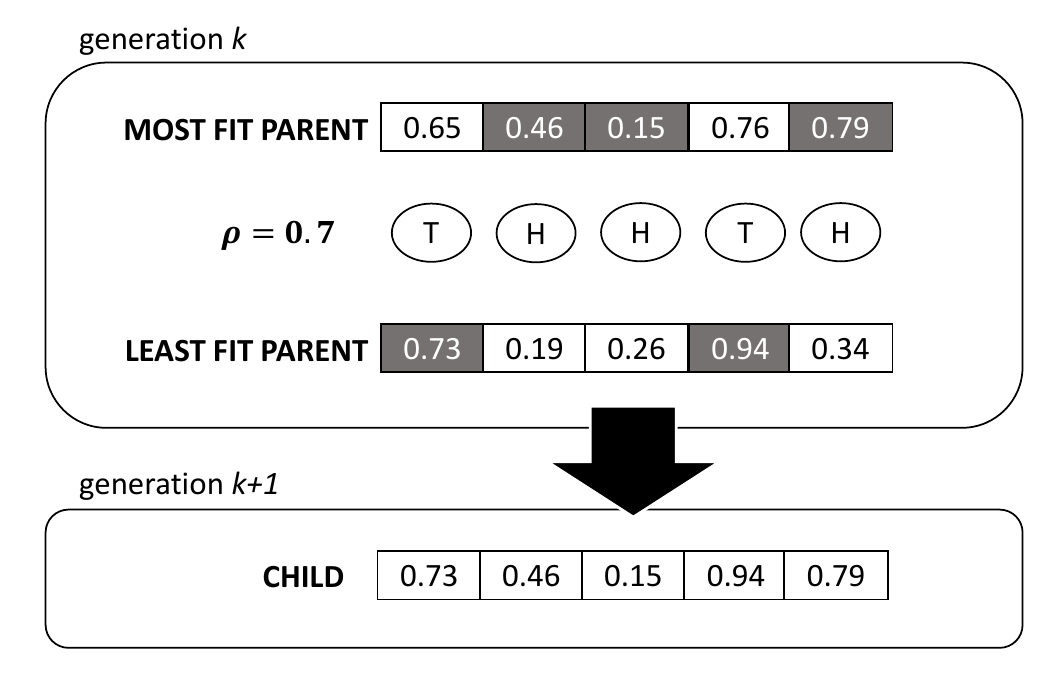}
    \caption{Mating process in BRKGA.}
    \label{Figure:Tutorial:Crossover}
\end{figure}

A critical aspect of a genetic algorithm is the combination of two solutions to generate offspring. In the basic BRKGA, as proposed by \citet{Goncalves2011:BRKGA}, each parent is a vector of random keys, and the mating process is performed using parametrized uniform crossover \citep{Spears1991:multi_point_crossover}. For each gene (each position in the random key vector), a biased coin is flipped to determine which parent will contribute its allele (key or gene value) to the offspring. The highest probability is assigned to the the most fit parent, prioritizing the selection of genetic material from the parent that represents a higher-quality solution. This bias towards the fitter parent contributes to the algorithm's tendency to preserve and propagate the characteristics of more promising solutions throughout the evolution process. Figure~\ref{Figure:Tutorial:Crossover} illustrates this crossover process where two parents with different fitnesses are combined. The most fit of the two will have a higher probability of passing on its genes to the child. A biased coin is flipped to determine which gene is passed on. An outcome of heads (H) corresponds to the most fit parents while tails (T), to the least fit.  

\subsection{The evolutionary process in the BRKGA}

The first generation, also known as the starting or initial population, consists of $p$ individual chromosomes, with each chromosome being composed of $n$ genes. Each gene is assigned a value (allele) that is generated uniformly at random within the interval $[0,1)$. To transform the random key sequence into a solution and simultaneously evaluate the solution's cost or fitness, each chromosome is processed by the decoder. Usually, decoded solutions are feasible, but infeasibility can be dealt with by penalties.

In generation $k$, the population is partitioned into two subsets: the elite and the non-elite. The elite subset comprises solutions with superior fitness scores, while the non-elite are the remaining solutions found in the population. Importantly, the size of the elite subset, determined by a parameter $p_e$ is intentionally smaller than that of the non-elite subset, prioritizing the preservation and evolution of high-performing solutions throughout the generations of the genetic algorithm.

To obtain generation $k+1$, as shown in Figure \ref{Figure:Tutorial:EvolutionaryProcess}, three operations are performed on the current population:

\begin{figure}[htb]
    \centering
    \includegraphics[scale = 0.45]{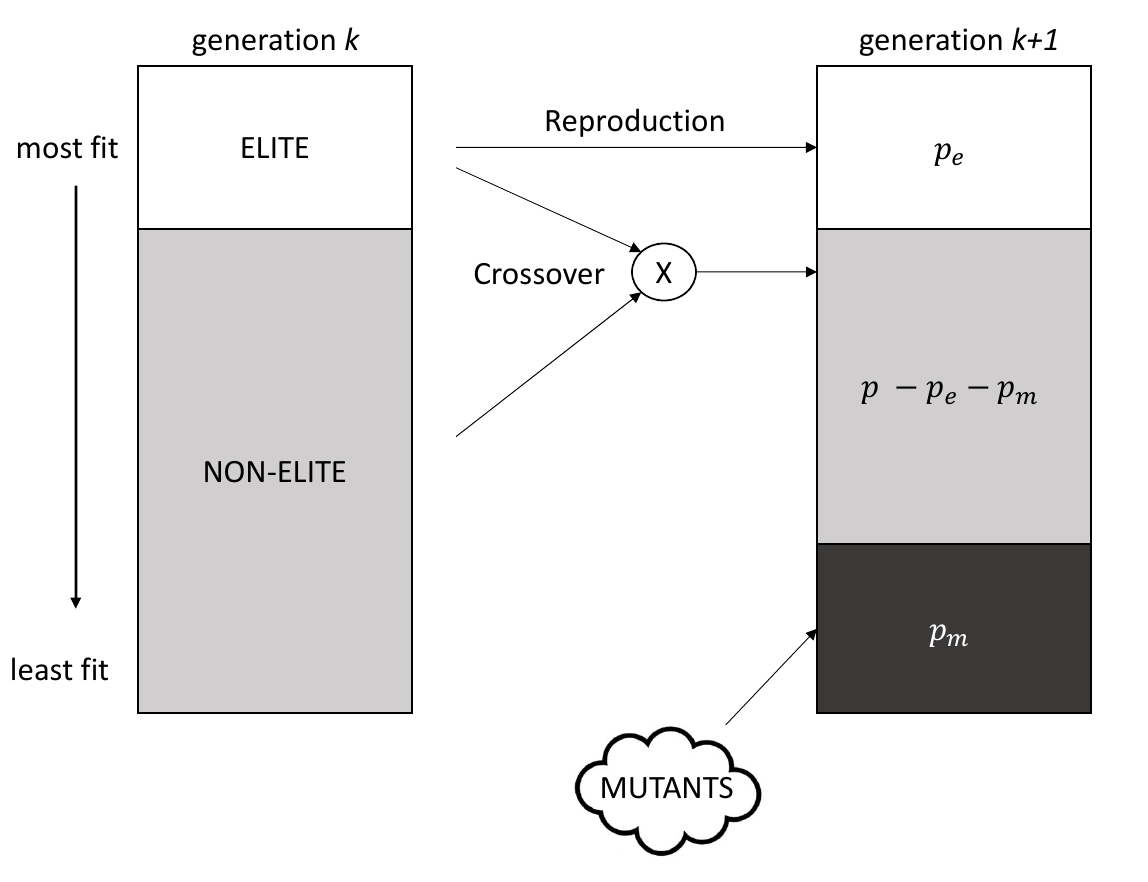}
    \caption{Evolutionary process between consecutive generations.}
    \label{Figure:Tutorial:EvolutionaryProcess}
\end{figure}

\emph{Reproduction} simply copies the $p_e$ chromosomes (random-key vectors) from the elite set of population $k$ to the successor population. This illustrates the elitist characteristic of the original BRKGA.

Creation of \emph{mutants} inserts $p_m$ new individuals in the successor population. These individual chromosomes are created as those of the initial population.

\emph{Crossover} performs the mating process as described in the Subsection~\ref{ss.tutorial.rk} to generate the remaining $p-p_e-p_m$ individuals in population $k+1$. To form the couple, a parent is randomly chosen from the elite set, while the other parent comes from the non-elite partition. In some papers, however, the second parent is selected from the entire population.

Once a new population is formed, its individuals are decoded and their costs are computed. Finally, the new population can be partitioned into elite and non-elite sets, and the evolutionary process restarts.

%------------------------------------------------------------%
%%%%%%%%%%%%%%%%%%%%%% Problems studied %%%%%%%%%%%%%%%%%%%%%%

\section{Main problems studied with BRKGA}
\label{Section:Applications}

To explore the areas of application of BRKGA and how they evolved, a thorough reading of~$156$ papers was performed. Throughout the literature review it was possible to identify ten categories of applications with at least four papers each, and~24 applications studied in up to three papers, for a total of~34 different applications. For illustration purposes, Figure~\ref{Figure:Results:Papers_problem_year} in the supplementary material shows the number of studies over the years per type of application domain, while Figure~\ref{Figure:Results:Papers_problem_category} presents the total amount of papers in each category.
Note that since we only consider indexed peer-reviewed papers in English from the Scopus database, the actual number of papers is probably greater than what we report. For example, we do not consider papers appearing in conferences nor in the arXiv repository.

In the following subsections, we present the applications of BRKGA, detailing the core ideas of these algorithms and how they evolved over time. The papers are presented in chronological order by micro-problems studied.

\subsection{Scheduling}

BRKGA has been extensively used for scheduling problems. In fact, almost a third of all BRKGA applications focus on this category of problems, which is by far the most studied problem with this metaheuristic. In 2002, \citet{Goncalves2002:hybrid_assembly_line_balancing} proposed a hybrid BRKGA with local search for the \textit{simple assembly line balancing problem}, which managed to outperform several algorithms from the literature on benchmark instances. The study from \citet{Moreira2012:assembly_line_worker_assignment_balancing} introduces the related problem \textit{assembly line worker assignment and balancing problem}. The proposed BRKGA+LS was compared favorably with a constructive strategy -- the same one used in the decoder -- and several algorithms from the literature. Later, \citet{Araujo2015:parallel_assembly_lines} study the \textit{parallel assembly line worker assignment and balancing problem}, an extension of the previous one. Their BRKGA outperformed a mathematical model but performed worse than a tabu search (TS) strategy.

Several papers study variants of the \textit{single batch processing machine problem} \citep{Wang2002:min_max_lateness_batch_processing,Kashan2006:single_batch_processing,Malve2007:max_min_lateness_parallel_batch_processing_dynamic_arrivals,Monch2018:matheuristic_batch_scheduling,Li2018:scheduling_bin_packing}. In \citet{Kong2020:parallel_scheduling_deterioration_maintenance}, the \textit{parallel batch scheduling with deterioration and learning effects on parallel machines problem} is introduced. The authors show that a hybrid BRKGA with a differential evolution algorithm has a better performance than both a particle swarm optimization algorithm and a traditional BRKGA. The same authors, in \citep{Kong2020:scheduling_uncertain_rolling_deterioration}, focus on the \textit{integrated steel production and batch delivery scheduling with uncertain rolling times and deterioration effect problem}. In this case, however, while a BRKGA outperformed dynamic programming and differential evolution approaches, it is outperformed itself by a VNS. 
\citet{Zhang2022:parallel_batch_scheduling_packing} studies the \textit{parallel batch processing machine scheduling problem under 2D bin-packing constraints}, which are added to increase applicability. An orthogonal BRKGA approach is developed, in which initial solutions are obtained with orthogonal design. This approach is shown to improve convergence and performance, with robuster and stabler results in comparison with several alternatives. \citet{Yu2023:supply_scheduling} investigates a \textit{two-echelon steel supply chain scheduling problem with parallel batch processing and deterioration effect}. A hybrid BRKGA with flower pollination operators is shown to be effective and efficient in comparison with alternatives such as PSO, FPA, and classic BRKGA.

The \textit{job shop scheduling problem} is another often explored with BRKGA \citep{Goncalves2005:hybrid_job_shop_scheduling,Goncalves2014:akers_graphical_jobshop_scheduling,Won2018:scheduling_cutting_parallel}. \citet{Homayouni2020:multistart_flexible_job_scheduling} studies the \textit{flexible job shop scheduling problem}. This work also used a population re-start operator, which acted similarly to a previous reset strategy in the BRKGA literature. This BRKGA outperformed a MILP and a number of approaches from the literature on benchmark instances. \citet{Fontes2023:bi_brkga_schedule_quay_cranes} explores a \textit{bi-objective energy--efficient job shop scheduling problem with transport resources}. The authors introduce a novel multi-objective approach to BRKGA, in which $\Pi + \Omega$ populations are evolved. Each of the $\Omega$ populations focuses on one of the objectives, while $\Pi$ populations consider all objectives and individual contribution to diversity. This alternative is shown to find Pareto Fronts close to the true Pareto Front, and outperforms NSGA-II in solution quality.

Meanwhile, \citet{Valente2006:early_tardy_scheduling} focuses on the \textit{single-machine scheduling problem with early and tardy penalties}. Their hybrid BRKGA+LS with a warm-start strategy outperformed many approaches from the literature, alongside variants without the warm-start and local search.  \citet{Valente2009:single_machine_scheduling} studies a variant of the previous problem, the \textit{single-machine scheduling problem with linear early and quadratic tardy penalties}. The proposed BRKGA with warm-start and local search had a better performance than several algorithms from the literature and variants without the warm-start and/or local search.

Four articles tackle variants of the \textit{project scheduling problem} \citep{Goncalves2008:resource_constrained_multi_project_scheduling,Mendes2009:resource_constrained_scheduling,Goncalves2011:foward_backward_improvement_constrained_scheduling}. In the most recent, \citet{Almeida2018:project_scheduling_flexible} study the \textit{resource constrained project scheduling with flexible resources problem}. Their decoder had a variable strategy: a gene indicates which of two algorithms is to be used to obtain the fitness value of a chromosome. This BRKGA outperformed constructive heuristics, and the BRKGA with variable decoding strategy was proved to be better than the ones with fixed strategies.

\citet{Goncalves2011:lot_sizing_scheduling_backorders} is the only study that applies BRKGA to the \textit{economic lot scheduling problem}. Their BRKGA was hybridized with an LP model and managed to outperform several approaches from the literature on randomly generated instances.
The next article studies the \textit{single-round divisible load scheduling problem}. \citet{Brandao2015:single_round_load_scheduling} propose a BRKGA that has a better performance than a multi-start algorithm combined with the decoding heuristic and some strategies used in the literature. Later, the same authors study the \textit{multi-round divisible load scheduling problem} in \citep{Brandao2017:scheduling_heterogeneous_systems}. Again in comparison with a multi-start algorithm and literature strategies, the proposed BRKGA was deemed the best approach.

The next two papers study the \textit{multi-user observation scheduling problem}. The first, authored by \citet{Tangpattanakul2015:multi_user_observation_scheduling}, compares BRKGA multi-objective variants, while the second, also by \citet{Tangpattanakul2015:scheduling_observations_satellite}, uses the hybrid decoder formulation suggested in the first. The latter study shows that an indicator-based multi-objective local search performed better than BRKGA for the studied problem.
\citet{Damm2016:field_tech_scheduling} focuses on the \textit{field technician scheduling problem}. Their BRKGA uses a novel elite diversification strategy that only lets individuals be copied in the reproduction step if they are significantly different from the other copied individuals. This was successful in preventing premature convergence and performed better than a MILP model and several constructive heuristics.
Later, the work from \citet{Cabo2018:bi_objective_scheduling} explores the \textit{bi-objective p-batch machine scheduling problem}. The BRKGA proposed considered the bi-objective nature of the problem in its decoding process and had competitive results with shorter computational time in comparison with two MILP models.

Two papers deal with the \textit{flowshop scheduling problem} \citep{Pessoa2018:flowshop_scheduling,Andrade2019:scheduling_software_cars}.  A generalization, the \textit{workforce allocation and two-stage flexible flow shop problem}, was studied in \citet{Bolsi2022:heuristic_workforce_allocation_scheduling}. BRKGA had the best performance on this lexicographic multi-objective problem in comparison with several heuristics, such as constructive methods, a random multi--start algorithm, a VNS, and a constraint programming model.
\citet{He2019:scheduling_setup_times} uses a hybrid BRKGA with an adaptive large neighborhood search algorithm to solve the \textit{order acceptance and scheduling problem}. Their approach outperformed many algorithms and MIP models from the literature on benchmark instances. The authors also introduced new instances for the problem.
The work from \citet{Andrade2019:scheduling_software_cars} introduces the \textit{time-- and machine--dependent scheduling problem}. In this work, both real-life-based and synthetic instances were used to prove the effectiveness of the BRKGA. In fact, the BRKGA had a better performance than several other algorithms, including ILS, simulated annealing, tabu search, genetic algorithm, and a MIP model.

\citet{Soares2020:scheduling_identical_parallel} aims to solve the \textit{identical parallel--machines problem with tooling constraints}. The authors use a hybrid BRKGA+VND+LS to solve benchmark instances and prove that their BRKGA performs better than both several approaches from the literature as well as a MIP model. \citet{Rocholl2021:scheduling_parallel_common_due_date} studies the \textit{parallel--machine multiple orders per job scheduling problem with a common due date}. The BRKGA was the best among the tested approaches, including a MILP and an ILS. The \textit{resource-constrained parallel--machine scheduling problem with setup times} is explored in \citep{Soares2022:resource_constrained_scheduling_setup}. Their BRKGA is hybridized with a VND with four neighborhoods, is shown to obtain tighter upper bounds for the problem in comparison with state-of-the-art methods. \citet{Maecker2023:unrelated_parallel_machine_scheduling} explores the \textit{unrelated parallel--machine scheduling problem with job--machine--dependent delivery times and eligibility constraints}. On randomly generated instances, the proposed BRKGA can outperforms several approaches from the literature, but needs higher computing time than a VNS to reach competitive solutions.

\citet{Abreu2021:scheduling_routing_makespan} focuses on the \textit{open shop scheduling problem with routing by capacitated single vehicle}, in which a scheduling problem is combined with a routing problem. Their hybrid BRKGA+IG approach outperformed a MIP model, a greedy insertion algorithm, an IGS, and a pure BRKGA on benchmark instances. This combination of problems is also explored in \citep{Kummer2022:home_health_care} on the \textit{home health care routing and scheduling problem}. The authors use BRKGA-MP-IPR \citep{Andrade2021:BRKGA_MP_IPR} and analyze the impact of each novel feature on solution quality and running times. The experiments indicate that BRKGA-MP-IPR with greedy cheapest-insertion constructive heuristic as decoder has better performance than state-of-the-art methods if longer running times are affordable.
\citet{Queiroga2021:ils_scheduling} presents a hybrid BRKGA with dynamic programming for the \textit{single machine total weighted tardiness batch scheduling problem}. When compared with a traditional ILS and a hybrid ILS+DP, the hybrid BRKGA outperformed its fellow hybrid, but both had worse performances than the classical ILS.
\citet{Silva-Soto2021:timetabling_bus_lines} study the \textit{bi-objective optimization approach for frequency setting and timetabling problem}. The authors compare single-- and bi--objective BRKGA with a MILP model. The bi--objective BRKGA used a non-dominated sorting algorithm in its decoder and proved to be the most effective approach for this problem.

The \textit{customer order scheduling problem with missing operations} was studied in \citet{Abreu2022:customer_order_scheduling}. The authors use a parameter--free restart procedure, that is performed when the worst fitness equals the best fitness in the population. A best improvement local search was also introduced. Their BRKGA outperformed several approaches from the literature on generated instances, with the combination of BRKGA+LS+restart being the best overall.
\citet{Xie2022:adaptive_brkga_cloud_workflow_scheduling} explores the \textit{cloud workflow scheduling problem}. An adaptive decoder is proposed in this paper, where one of three decoding methods is chosen by a random probability. This probability changes during evolution, decreasing chance of the more complex decoders and increasing the chance of a simpler decoding strategy. A warm-start strategy is employed, where some initial chromosomes come from heuristics based on list scheduling and the others are based on a level heuristic. A local search is also customized. The experiments show that adaptive decoding BRKGA has a robust performance and outperforms several approaches from the literature.
\citet{Fontes2023:bi_brkga_schedule_quay_cranes} investigates \textit{joint scheduling quay cranes and speed adjustable vehicles in container terminals}. The bi-objective BRKGA uses the same strategy as \citet{Fontes2023:job_shop_transp_resources}. Experiments shown that the proposed BRKGA obtains diverse and uniformly distributed solutions, which cover, almost always, the whole Pareto Frontier of the instance.
Finally, \citet{Zhao2023:distributed_sequential_assembly_process} uses a floating point number encoding to solve the \textit{multi--station multi--robot task allocation and sequential planning problem}. In the chromosome, the digit before ``.'' indicates the work station, and the digits after, the robot allocated to the job. Due to problem characteristics, the authors also introduce three novel evolution operators. Double crossover uses two operations, in which the digits before and after ``.'' are analyzed independently. The same is done with double mutation. The elite re-optimization operator diversifies the elite set if the minimum cost of the current and next generations are close by applying classic crossover and mutant operators. Experiments show that the proposed method outperforms a greedy insertion algorithm and a hybrid BRKGA+SA from the literature.

Considering the many studies in this category, it is difficult to identify common trends for scheduling problems. There is approximately the same amount of studies that either use benchmark or randomly generated instances, with the latter being a little more popular, and the slight majority of papers use permutation-based decoders. However, one trend common to most studies is the use of hybridization, generally with local search algorithms, whose positive impact on performance makes BRKGA the best approach to many of the studied problems.

\subsection{Network configuration}

Several papers focus on the \textit{weight setting problem in OSPF routing}. The first, \citet{Ericsson2002:Genetic_alg_OSPF}, introduces a BRKGA with warm-start that performed better than several state-of-art algorithms. This BRKGA was later extended by \citet{Buriol2005:weight_setting_problem_OSPF_routing}, who added a local search operator to the previous framework. This addition outperformed both the previous BRKGA and the pure local search. \citet{Buriol2007:OSPF_routing} changes the fitness function of their previously used BRKGA and observes which fitness calculation leads to better solution quality. Finally, \citet{Reis2011:OSPF_DEFT_routing_network_congestion} extends the BRKGA of \citet{Ericsson2002:Genetic_alg_OSPF} and \citet{Buriol2005:weight_setting_problem_OSPF_routing} for \textit{DEFT routing}. This BRKGA was outperformed by a two-stage approach.

\citet{Fontes2007:concave_minimum_cost_network_flow} studies the \textit{single source uncapacitated concave minimum cost network flow problem}. Their hybrid BRKGA with local search managed to outperform a pure BRKGA on benchmark instances. A related problem, the \textit{hop-constrained minimum cost flow spanning tree problem}, was studied by \citet{Fontes2013:multi_pop_hop_constrained_trees_nonlinear}. Their hybrid BRKGA+LS with a multi--population strategy found the best solutions for all instances and had better results in comparison with a mathematical model.

The \textit{survivable IP/MPLS--over--WSON multilayer network optimization problem} was explored in two papers. \citet{Ruiz2011:WSON_multilayer_network} introduced a BRKGA and a mathematical model for this problem, while \citet{Pedrola2013:grasp_pr_survivable_network} compared a multi--population BRKGA with several GRASP approaches. The authors of the latter showed that a hybrid GRASP+PR approach was the most effective.

\citet{Noronha2011:routing_wavelenght_assignment} studies the \textit{routing and wavelength assignment problem}. The proposed BRKGA had a better performance when compared with a state-of-the-art tabu search algorithm and other approaches from the literature. Later, \citet{Brandao2016:accepted_lightpaths_networks} focused on the same problem. The authors used two different algorithms as decoders, and show that the resulting BRKGAs are more effective than many constructive algorithms in the literature. Finally, \citet{Pinto2020:routing_assignment} compared a BRKGA with the best algorithms in the literature and a multi--start version of the decoder algorithm, in which the BRKGA was considered the most effective.

\citet{Velasco2013:capex-flexgrid_optical_networks} focuses on three strongly related problems: the \textit{area partitioning problem}, the \textit{IP/MPLS area network design problem}, and the \textit{core network design problem}. The authors introduce three different BRKGAs, each focusing on one of those problems, and observe the quality of the given solutions in close-to-real instances. 
The focus of \citet{Pedrola2013:hybridizations_regenerator_placement_dimensioning} is the \textit{regenerator placement and dimensioning problem}. The authors hybridize a BRKGA with both VND and path relinking. This hybrid algorithm outperformed several approaches, including a MILP and an evolutionary multi-start GRASP+VND+PR.

\citet{Moran-Mirabal2013:handover_minimization_networks} studies the \textit{handover minimization problem}. The proposed BRKGA+LS hybrid showed better results in comparison with a MIP model but had worse results than an evolutionary GRASP with path relinking.
\citet{Andrade2015:wireless_backhaul} tackles the \textit{wireless backhaul network design problem}. In real-life inspired instances, the proposed BRKGA outperformed both a MIP model and a multi-start algorithm.
\citet{Ruiz2015:capacitated_minimum_spanning_tree} focuses on the \textit{capacitated minimum spanning tree problem}. The authors introduce different decoders, alongside a local search strategy, strategic oscillation to explore new areas of the solution space, and a neighborhood reduction strategy to ignore non-feasible moves on the tree. In comparison with other alternatives, the BRKGA had higher solution quality and lower computational effort, especially when combined with the local search algorithm.

Three articles focus on the \textit{distribution network reconfiguration problem} \citep{Cavalheiro2018:power_distribution_networks,deFaria2017:distribution_network,Raposo2020:network_reconfiguration_energy_loss}.In the most recent paper, \citet{Raposo2020:network_reconfiguration_energy_loss} adapts the BRKGA for multi--objective optimization and compares it favorably with a NSGA-II on benchmark instances.
Finally, the paper authored by \citet{Mikulski2021:energy_storage} uses a multi--objective BRKGA to \textit{reduce transmission losses in power distribution networks}. This multi--objective approach found better results than a multi-objective particle swarm optimization (PSO) and a NSGA-II on four test cases.

For network configuration problems, the use of indicator-based decoders is of note, something done by more than half of the studies. Another characteristic in common is the presence of hybridization, with either local search, warm start, or multi-population strategies. The use of benchmarks is common, as it happens on almost three-fourths of the papers in this category.

\subsection{Location}

In this subsection, problems that consider the location of one or several facilities are considered. One such problem is the \textit{tollbooth problem}, whose BRKGA was proposed in \citet{Buriol2010:road_congestion} and later extended in \citet{Stefanello2017:traffic_congestion_tolls}. Those studies show that the approach with local search performs significantly better in time and quality for benchmark instances, in comparison with exact models and a local search-free approach. Another such problem is the \textit{regenerator location problem} studied in \citet{Duarte2014:regenerator_location}. Using benchmark instances, the authors show that BRKGA has a better performance than the ones in the literature, but a GRASP+LS hybrid outperforms it.

A version of the \textit{quadratic assignment problem}, the \textit{unequal area facility layout problem}, is studied by \citet{Goncalves2015:unequal_area_facility_layout}. The proposed approach hybridizes BRKGA with linear programming constraints. The hybridized version outperformed the pure BRKGA and several other algorithms from the literature. In fact, both BRKGA versions improved the best-known solution of several benchmark instances. \citet{Lalla-Ruiz2016:quadratic_assignment} studies the \textit{quadratic assignment problem}. Their BRKGA has an improvement phase, similar to a local search, and performs better than a migrating birds algorithm and a discrete differential evolutionary algorithm with a local search portrayed in literature. Similarly, \citet{Stefanello2019:placement_virtual_machines} studies both a generalization of the previous problem and its specific application. This paper also hybridizes a BRKGA with path-relinking and local search. This hybrid BRKGA outperforms mathematical models and a hybrid GRASP-PR-LS for smaller, randomly generated instances.

Both \citet{Lopes2016:hub_location_routing} and \citet{Pessoa2017:tree_hubs_location} study \textit{hub-location problems}. The approach in \citet{Lopes2016:hub_location_routing} uses a hybrid BRKGA with local search and warm-start, which performs poorly in comparison with multi-start VND and local search, and an integer programming model with valid inequalities. Meanwhile, \citet{Pessoa2017:tree_hubs_location} compares several BRKGAs and proves this approach's effectiveness and efficiency. A \textit{two‑level hub location routing problem with directed tours} is studied in \citet{Freitas2023:brkga_two_lvl_hub} with a modified decoding process. Their BRKGA outperformed a MIP and the state-of-the-art variable neighborhood decomposition search and improved the objective value for many of the benchmarks from the literature.

The study of \citet{Biajoli2019:capacitated_facility_location} focuses on the \textit{two-stage capacitated facility location problem}. Their BRKGA+LS algorithm outperforms the genetic algorithm and clustering search presented in the literature on known benchmark instances, and it is extended to a \textit{multi-product} version in \citet{Mauri2021:multiproduct_facility_location}. This extended algorithm also manages to outperform a clustering search algorithm and several mathematical models. 

\citet{Johnson2020:cover_by_pairs} studies two special cases of the \textit{cover-by-pairs optimization problem}. In experiments with synthetic and real-world-based problems, the BRKGA outperforms its only competitor in the \textit{path-disjoint} case, a multi--start greedy algorithm. However, in the \textit{set-disjoint} case, a heuristic-based approximation algorithm is the winner.
\citet{Londe2021:pnext_center} uses a BRKGA-MP-IPR~\citep{Andrade2021:BRKGA_MP_IPR} with shaking, warm-start, and local search to solve the \textit{p-next center problem}. The authors show that their algorithm has a better performance than the best in the literature, which was a GRASP+VND hybrid.
Finally, \citet{Villicana2022:mobile_labs_covid19_testing} studies the \textit{accessibility location problem for COVID-19 test sites.} The proposed approach has the objective of increasing accessibility. BRKGA obtained solutions with similar quality to a MILP model with a fraction of its computational time.

In general, the adaptations of BRKGA to location problems have some common characteristics. All studies use permutation-based decoders, i.e., the fitness evaluation of the chromosomes involves the sorting of the random keys, except \citep{Johnson2020:cover_by_pairs}, which uses an indicator-based decoder, and \citep{Freitas2023:brkga_two_lvl_hub,Villicana2022:mobile_labs_covid19_testing}, whose approaches combine indicator and permutation methods. The great majority of applications also hybridize BRKGA with local search algorithms and use benchmark instances to observe algorithm performance. Lastly, BRKGA tends to outperform or match the best approaches in the literature for problems in this category.

\subsection{Cutting and Packing}

One of the first applications of the defined BRKGA framework was  \citet{Goncalves2011:multi_pop_constrained_2d_orthogonal_packing}, which explores the \textit{container loading problem}. The hybrid BRKGA with multi-populations was shown to perform better than several alternatives from the literature, including GRASP, VNS, genetic algorithms, and tabu search algorithms. Meanwhile, \citet{Goncalves2013:2d_3d_bin_packing} studies the \textit{2D and 3D bin packing problems}. Their approach statistically outperforms the ones presented in the literature, which include tabu search, local search, and GRASP+VND approaches. The \textit{3D bin packing problem} was also studied by \citet{Zudio2018:BRKGA_VND_bin_packing}, which hybridizes a BRKGA with a VND. This combination was shown to be effective in comparison with pure BRKGA, with better solutions in fewer generations.

Meanwhile, two papers focus on variants of the \textit{cutting and packing problem}. \citet{Mundim2017:open_dimension_nesting} first introduces a BRKGA approach, which is then compared favorably with the results of GRASP and MIP models from the literature on benchmark instances. Later, \citet{AmaroJunior2017:multiple_pop_irregular_strip_packing} uses a different strategy in its decoding phase, and either outperforms or matches results found in the literature.
\citet{SouzaQueiroz2020:irregular_knapsack} studies the \textit{2D cutting problem with irregular shaped items}. The authors observe that the BRKGA had solutions of similar quality to the ones from a general VNS. However, the BRKGA was significantly faster than its competitor. Similarly, the \textit{additive manufacturing production planning problem}, which is a version of the 2D irregular packing problem, was explored in \citet{Lu2023:additive_manufacturing_production}. A BRKGA is at the heart of the pixel-based AM packing algorithm (PAMPA). which can be used to check solution feasibility.

Later, \citet{Goncalves2020:two_dimensional_cutting_defects} introduces a BRKGA for the \textit{2D non-guillotine cutting problem}. Using benchmarks from the guillotine version, the proposed algorithm is faster and has higher solution quality than two of the best algorithms in the literature. The same problem was studied in \citet{Oliveira2022:two_dimensional_non_guillotine_cutting}, with an approach based on the one from \citet{Goncalves2013:2d_3d_bin_packing}. The proposed BRKGA adds a procedure to group plates into blocks. This addition improves performance, with results equal to or better than those previously published.
\citet{AmaroJunior2021:minimum_time_cut_path} uses a BRKGA for the \textit{minimum time cut path problem}. This BRKGA uses a decoder that gives both cut order and direction. The proposed approach outperformed a GA and a commercial software on artificially generated instances.
Lastly, \citet{Oliveira2022:diversity_scenario_gen} uses a \textit{packing problem with uncertainty} to evaluate performance of a scenario-generating BRKGA. In it, each chromosome represents a possible scenario, which is evaluated regarding diversity and impact on a given first-stage solution.

For packing problems, the BRKGA was mostly used in similar ways. In fact, all proposed BRKGA except \cite{Zudio2018:BRKGA_VND_bin_packing} use permutation-based decoders without any sort of hybridization. Not only that, but almost all articles use benchmark instances, and in all studies, BRKGA was considered the best algorithm with respect to solution quality and computational times.

\subsection{Vehicle Routing}

\citet{Grasas2014:blood_collection_clinical} explores the \textit{blood sample collection problem}. The BRKGA was shown to be an interesting alternative, in comparison with a MIP model, that could be easily used in the practical applications of this study.
Three papers focus on \textit{uni-- and multi--directional road network problems} \citep{Huang2018:road_network_disruptions,Huang2020:model_urban_road_network_disruptions}. In the most recent one, \citet{Huang2020:bi_objective_road_disruptions} adds disruption constraints to the problem. The proposed BRKGA was outperformed by an ILS.

\citet{Ruiz2019:routing_capacity_distance} studies the \textit{open vehicle routing problem}. The authors combine a BRKGA with a local search and a strategic oscillation operator. In comparison with several algorithms in the literature and a GRASP, the hybrid BRKGA had the best performance. The work of \citet{Ibarra-Rojas2021:routing_equity} focuses on a generalization, the \textit{vehicle routing problem with egalitarian distribution}. The authors compare three decoders and introduce a mathematical model for the problem.
\citet{Carrabs2021:set_orienteering} studies the \textit{set orienteering problem}. Their BRKGA was hybridized with three simultaneous local searches, a reset operator, and a multi-population strategy. This approach was shown to be highly effective in comparison with a VNS and a matheuristic.
\citet{Schenekemberg2022:dial_a_ride} focuses on the \textit{dial-a-ride problem with private fleet and common carrier}. To solve this problem, the authors use BRKGA-QL, a hybridization with Q-Learning-based parameter control. They also propose a local search with seven neighborhoods. The hybrid BRKGA-QL+LS dominated a BRKGA-QL and a branch and cut algorithm in performance and computational time on both benchmark instances and a case study.

\citet{Schuetz2022:robot_traject_planning} explores \textit{robot-trajectory planning problems}. A novel BRKGA-based scheme is proposed, where several methods can create solutions for the first generation of the evolution. BRKGA was shown to be faster than a SA applied within a Quadratic Unconstrained Binary Optimization approach but lost in speed to a GA with an SA-based evolution operator.
Finally, \citet{Marques2023:two_phase_multi_obj_UAV_routing} studies the \textit{multi-objective green routing drone grid problem}, where the airspace is divided into horizontal and vertical bands. BRKGA was shown to obtain viable solutions quickly but was outperformed by VND-based methods.

There are some trends common to BRKGA used to solve routing problems. The use of real-life-based instances is a characteristic of this category, with almost half of the studies doing such. Another common trend is the use of permutation-based decoders, done by most articles. Lastly, BRKGA was shown to be the better algorithm on almost all papers, except those in which it is compared with ILS algorithms. 

\subsection{Traveling salesman problem}

\citet{Moran-Mirabal2014:family_traveling_salesperson} is the first use of BRKGA for any generalization of the TSP. Specifically, the authors study the \textit{family traveling salesman problem}. Their BRKGA approach shows better results than a MIP model and a hybrid GRASP+PR heuristic on smaller instances, but the hybrid GRASP was shown to be better on the larger instances. The same problem is studied in \citep{Chaves2024:adaptive_ftsp}, with a chromosome whose last gene indicates which of the five decoding procedures will be used to decode the individual. The BRKGA approach uses Q-Leaning to control its parameters during evolution, alongside a random VND to control a local search with six neighborhoods, and a perturbation component for the crossover between similar individuals. The Q-Learning approach outperformed state-of-the-art algorithms, including the one from \citet{Moran-Mirabal2014:family_traveling_salesperson}, and a parallel branch-and-cut algorithm on larger instances.
Later, \citet{Bernardino2018:traveling_purchaser} addresses the \textit{traveling purchaser problem}. For this, the authors propose two different BRKGA strategies, both hybridized with a local search. In comparison with a hybrid GA+LS, there was no difference in solution quality among the three algorithms, but the hybrid genetic algorithm had considerably smaller computational times.		

\citet{Silva2019:multicommodity_tsp} study the \textit{multi-commodity TSP with priority prizes}. The authors propose two hybrid BRKGA with ILS, where the ILS itself is hybridized with a VND. One of the BRKGA approaches is an adaptive BRKGA (A--BRKGA), introduced in \citet{Chaves2018:capacitated_centered_clustering_local_search}. In the experiments, the A--BRKGA+ILS was not shown to be statistically better than the BRKGA+ILS, but both had better performances than a MIP model. 

The work of \citet{Chagas2020:traveling_salesman_loading} introduces the \textit{double TSP with partial last-in-first-out loading constraints}. Their BRKGA performs better than two LP models on the proposed instances. Finally, \citet{Chagas2021:bi_objective_traveling_thief} focuses on the \textit{bi-objective traveling thief problem}, in which the TSP is combined with a knapsack problem. The bi-objective approach hybridizes a BRKGA with an NSGA-II, alongside warm-start and a local search algorithm. This BRKGA either matched or outperformed several approaches in the literature.
\citet{Junior2023:a_brkga_laser_cutting} studies a generalization of the TSP called \textit{laser cutting path planning problem}. In it, the objective is to minimize the time the laser takes between the corners of a cutting form. The authors hybridize A--BRKGA with a perturbation strategy. A Eulerian path heuristic is used to generate initial solutions. The BRKGA approaches outperform a MIP model, though A--BRKGA had worse performance than its classic counterpart.

The studies in this category have great diversity in the choice of instances, with artificially generated, new, and benchmarks being used. This diversity is not extended to BRKGA performance in comparison with other algorithms, which follow a specific trend: namely, BRKGA outperforms most exact approaches, but tends not to be effective with respect to other metaheuristic algorithms. Also in common is the use of permutation-based decoders, as all papers in this category of problems apply.

\subsection{Clustering}

\citet{Festa2013:data_clustering} uses a BRKGA for clustering biological data. This BRKGA uses an improvement procedure inside its decoder that is similar to a local search approach. Compared with several state-of-the-art algorithms, including variations of hybrid GRASP+PR, the BRKGA was effective and efficient in finding good solutions. This BRKGA was adapted by \citet{Oliveira2017:hybrid_constrained_clustering} for the \textit{constrained clustering problem}. The authors also add to the algorithm a local search external to the decoding process. In this case, the BRKGA+LS was outperformed by a column generation algorithm hybridized with PR and local search.

\citet{Chaves2018:capacitated_centered_clustering_local_search} introduces the adaptive BRKGA with clustering search for the \textit{capacitated centered clustering problem}. The A--BRKGA is a variation that employs on--line parameter tuning. Its combination with clustering search was shown to have the best performance compared to a classic BRKGA and other state-of-the-art algorithms. 
This problem is also explored with A--BRKGA in \citet{Xu2022:ABKRGA_local_search}, but with an iterative neighborhood local search. The novel approach is shown to improve known solutions with a small increase in running times.

\citet{Martarelli2020:feature_selection} uses BRKGA for \textit{unsupervised feature selection}. The authors introduce two variations of BRKGA. The difference between the two was the use of warm-start. BRKGA was the best approach compared with others in the literature, but the inclusion of warm-start solutions led to better results.
The paper by \citet{Fadel2021:statistical_disclosure_control} studies the \textit{multivariate micro aggregation problem}, in the context of statistical disclosure control. The authors introduce two decoders for this problem. The proposed BRKGA was compared with several other methods and consistently found better results than its competitors.
Finally, \citet{Brito2020:stratified_sampling} proposes a BRKGA for the \textit{k-medoids clustering problem}. The authors detail a crossover operator that works as a local search with best improvement strategy. Tests show the efficacy of the approach in comparison with several from the literature.

There are some characteristics in common in the papers that focus on clustering. Most of the article use indicator-based decoders. All articles test their proposed approaches on benchmarks or known datasets, and most of them hybridize BRKGA with local or clustering search algorithms. 

\subsection{Graph problems}

Clique problems are the focus of several articles \citep{Fontes2018:maximum_edge_weight_clique,Pinto2018:maximum_quasi_clique,Pinto2021:maximum_quasi_clique_local_search}. In the most recent one, \citet{Melo2023:brkga_quasi_clique_partitioning} studies the \textit{minimum quasi-clique partitioning problem}. Their approach was shown to obtain results at least as good as the ones in the literature and to find better results for a fifth of the instances.

\citet{Lima2022:matheuristic_broadcast_time} focuses on the \textit{minimum broadcast time problem}. The proposed approach combines a matheuristic-based BRKGA with two possible decoders. Experiments show that the matheuristic outperformed state-of-the-art methods in quality and computational time.
\citet{Silva2023:brkga_chordal_completion} tackles the \textit{chordal completion problem}. The proposed BRKGA uses a permutation-based decoder to eliminate possible orderings of triangulation in the graph. In a comparison with the winner algorithm of the Parameterized Algorithms and Computational Experiments Challenge 2017, BRKGA was shown to find feasible solutions for all instances in which the model failed and to improve solution quality.
Finally, \citet{Londe2022:root_sequence_index} introduces the \textit{root sequence index allocation problem}. This telecommunication problem is tackled as a generalization of the classical vertex coloring problem. The authors customize three decoding approaches, two local searches, a warm-start procedure, and shake and reset operators for two versions of this problem, with differing objectives. BRKGA was shown to obtain better solutions than an ILS and VNS on one of the objectives and to be comparable to VNS on the other.

For this category of problem, all BRKGAs use permutation-based decoders on benchmark instances except \citep{Londe2022:root_sequence_index}, which proposes an indicator-based decoder and introduces new instances. Also noteworthy is the quality of the obtained solutions, with BRKGA consistently outperforming several state-of-the-art algorithms.

\subsection{Parameter optimization}

In \citet{Caserta2016:data_fine_tuning}, the BRKGA is used to tune a cross entropy-based scheme, itself used to solve a binary classification problem. \citet{ErzurumCicek2021:artificial_network_time_series} uses a BRKGA to \textit{optimize ANN parameters with time series forecasting}. The algorithm was successful with respect to several classical approaches, such as SARIMA, SVR, ARIMA, and classical GA, but had a clear disadvantage in computational times.

The work of \citep{Sun2022:bayesian_network_structure} focuses on \textit{bayesian network structure learning}. The authors use the algorithm NOTEARS as both a decoder and local optimizer to solve the non-convex problem. The proposed BRKGA can achieve a good performance on benchmark networks and on a real data set in comparison with pure NOTEARS, classical GA, and hybrid PSO+GA.
\citet{Falls2022:ocean_model_parameter_optimization} uses BRKGA to \textit{estimate the ideal set of parameters} to simulate the behavior of the biogeochemical component of an ocean model. In experiments, the authors observe the ability of BRKGA to create simulations and compare those with real data.
Finally, \citet{Japa2023:hybrid_brkga_hyperparameter_neural_nets} optimizes \textit{hyperparameters for neural networks}. BRKGA+Bayesian Walk exploitation procedure was shown to produce better results in a more consistent manner than several approaches from the literature.

Some conclusions may be observed for the papers in this category of application, such as that all proposed BRKGA approaches use indicator-based decoders. One point that is always observed is the speed of the algorithm in comparison with others from the literature to solve both benchmarks and real-life datasets. It is also of note that BRKGA tends to outperform other metaheuristic approaches for problems in this category.

\subsection{Container loading}

\citet{Goncalves2012:multi_pop_container_loading} introduces a multi-population BRKGA for the \textit{3D container loading problem}. This algorithm had the best overall performance in comparison with several other approaches, including genetic algorithm, branch-and-bound, tabu search, and GRASP. Later, \citet{Zheng2015:multi_objective_container_loading} studies the same problem, also with a multi-population BRKGA, but with a bi--objective spin. The bi--objective approach outperformed several variants on benchmark instances.

\citet{Ramos2016:container_loading} hybridizes a multi-population BRKGA with a constructive heuristic, an approach shown to be effective in comparison with several other algorithms. These algorithms include the one in \citet{Goncalves2012:multi_pop_container_loading}. This multi-population BRKGA was later extended by \citet{Ramos2018:container_loading} to consider other necessary constraints, leading to improvements in solution quality and efficiency.

Even though the container loading problems are very similar, in concept, to packing problems, the authors of the former problem use different strategies than those of the latter. In this category, it is of note the extension of already known and well-performing BRKGA approaches to consider new problem characteristics. In fact, the comparison of performance among proposed algorithms is frequent in this category, alongside the use of known benchmarks, permutation-based decoders, and multi-population strategies.

\subsection{Other problems}

This subsection details studies that focus on problems that are not included in any of the previous categories. The first article, \citet{Goncalves2004:manufacturing_cell_formation}, uses a hybrid BRKGA+LS for \textit{manufacturing cell formation}.  \citet{Resende2012:steiner_triple_covering} studies the \textit{Steiner triple covering problem}. Another application of BRKGA presented by \citet{Silva2014:roots_non_linear_equations} is to solve \textit{non-linear systems of equations with multiple roots}. 
Two studies focus on \textit{berth allocation problems}. The first, \citet{Lalla-Ruiz2014:tatical_berth_allocation}, outperformed both a tabu search and several branch-and-price approaches. Later, \citet{Correcher2017:berth_allocation} hybridizes a BRKGA with two distinct local searches and a matheuristic. 

Similarly, the \textit{unit commitment problem} is studied in two papers. \citet{Roque2014:unit_commitment} introduces a BRKGA+LS hybrid heuristic. \citet{Roque2017:multi_objective_unit_commitment} studies a multi-objective version of the previously mentioned problem. The authors use the same decoder as in the previous study, but in a multi-objective BRKGA. 
\citet{Andrade2015:winner_auctions} focuses on the \textit{winner determination problem in combinatorial auctions}. The authors use a BRKGA with warm-start solutions originating from a LP model, alongside three decoding strategies. \citet{Chan2015:inventory_multi_item_lot_sizing} tackles the \textit{multi-item capacitated lot sizing problem}. The \textit{cloud resource management problem} is the focus of \citet{Heilig2016:resource_management_cloud}.

\citet{Goncalves2016:open_stacks} studies the \textit{minimization of open stacks problem}.
\citet{Chaves2016:hybrid_tool_switches} tackles the \textit{minimization of tool switches problem}. The proposed approach hybridizes BRKGA with a clustering search algorithm. 
\citet{Hottung2016:container_pre_marshalling} focuses on the \textit{container pre-marshaling problem}. \citet{Andrade2017:feasible_mixed_integer} hybridizes a BRKGA with a feasibility pump to find \textit{feasible solutions to MIP}. The \textit{protein--ligand flexible molecular docking problem} is studied by \citet{Leonhart2019:protein_docking}. 

\citet{Oliveira2018:car_rental} study the \textit{car rental problem}. The proposed methodology includes a warm-start with constraint programming and integer non-linear programming models, linear programming constraints included in the decoding process, and a MIP model applied to the best solution found by BRKGA. \citet{Oliveira2019:matheuristic_car_rental_stochastic} explores the \textit{car rental capacity-pricing stochastic problem}. The authors combine a BRKGA with a matheuristic to generate solutions for a two-stage stochastic problem and show the efficiency of the proposed approach in several instances. The study of \citet{Brito2020:stratified_sampling} focuses on \textit{optimal allocation in stratified sampling}. \citet{Pinacho-Davidson2020:minimum_capacitated_dominationg_set} tackles the \textit{minimum capacitated dominating set problem}. The proposed BRKGA uses an  ILP solver inside its decoder and outperforms a state-of-the-art local search heuristic and a MIP model.

\citet{Pan2021:covid19_spread} uses BRKGA in the decision-making phase of a framework to \textit{mitigate Covid-19 spread}.  \citet{Ochoa2021:search_trajectory_networks} compares the \textit{search trajectory network} of several metaheuristics. BRKGA is used in the combinatorial optimization case study, in which a $p$-median problem was explored. The authors observe that BRKGA quickly gets trapped in different areas of the solution space and visits a higher number of solutions compared to other metaheuristics.
The work of \citet{Pastore2022:bezier_brkga_topology} studies \textit{topology optimization for stress-constrained structures}. This problem is in the context of additive manufacturing, and the authors use B\'ezier curves to obtain the best paths. To test the results, the authors made a reduced-scale print of a piece and observed its structural capabilities. \citet{Silva2023:BRKGA_maximum_diversity} explores the \textit{maximum diversity problem}. Experimental results in benchmarks indicate that a hybrid BRKGA+LS outperforms other approaches from the literature in terms of solution quality. Finally, \citet{Morgan2023:multispacecraft_maneuvers} searches for \textit{multi-spacecraft maneuvers for mobile target tracking}. Two case studies point out that a BRKGA-based method outperforms its competitors but with diminishing returns for greater spacecraft numbers.

\subsection{Summary}

The studies detailed in this section highlight several points in regard to BRKGA use and characteristics. First, the decoders are frequently permutation-based, i.e., use sorted random-keys as a part of the decoding process. Second, the use of hybridization is noted to improve BRKGA performance, especially when a local search or matheuristics are used. Third, the majority of papers use benchmark instances, and compare the proposed algorithms with state-of-the-art approaches. Fourth, BRKGA almost always outperforms MIP models and tends to obtain similar or better results than other algorithms. Fifth and last, authors do not frequently build on already existing BRKGA approaches, instead introducing new decoding processes or operators.

The scenario shown in this section can be summarized as such: BRKGA is a versatile algorithm, used successfully in several problems, and that improves significantly when hybridized. The use of hybridization is, thus, crucial for an efficient and well-performing BRKGA, and is the focus of the next section.

%------------------------------------------------------------%
%%%%%%%%%%%%%%%%%%%%%%% Hybridizations %%%%%%%%%%%%%%%%%%%%%%%

\section{Main hybridizations with BRKGA}
\label{Section:Hybrids}

To observe the evolution of hybridization with BRKGA, another detailed reading of the selected papers was performed. In this case, the objective was to observe and detail other algorithms or frameworks simultaneously used with BRKGA to solve different optimization problems. We explore 94 papers that use diverse hybrid strategies. For illustrative purposes, the number of studies over the years can be seen in Figure~\ref{Figure:Results:Papers_hybrids_year}, while Figure~\ref{Figure:Results:Papers_hybrids_category} shows the total amount of papers in each category. Please note that papers can be included in more than one category of hybrid.

\subsection{Local search heuristics}

The complementary use of two or several heuristics to explore all available knowledge is the basis of memetic algorithms~\citep{Moscato1989:memetic}. The use of local search (LS) algorithms as the intensification phase of an evolutionary algorithm is a well-known application of the memetic framework, and tries to mimic the idea of Lamarck evolution, that individuals may pass characteristics obtained during their lives to their descendants~\citep{Cotta2018:memetic_alg}. Several of the papers that apply local search alongside BRKGA do not give extensive details about LS neighborhood or stopping criteria, or when it is applied in the main loop of the algorithm. Details on the complexity of the LS are also, in general, scarce. Thus, in the first part of this section, the use of simpler local searches -- not based on specific heuristic frameworks -- is addressed. They are divided by which module of the BRKGA the LS is embedded in and by the neighborhood explored. The second part of this section explores the application of known heuristics as the LS of BRKGA.

In BRKGA, the use of local search algorithms may be divided into two categories, indicating in which phase of the GA it is used. The most frequent approach is the use inside the decoding phase, either before, e.g., 
\citet{Festa2013:data_clustering}\,
\citet{Fontes2007:concave_minimum_cost_network_flow,
Fontes2013:multi_pop_hop_constrained_trees_nonlinear},
\citet{Goncalves2014:akers_graphical_jobshop_scheduling,Goncalves2011:foward_backward_improvement_constrained_scheduling},
\citet{Heilig2016:resource_management_cloud},
\citet{Moran-Mirabal2013:handover_minimization_networks}, 
\citet{Ruiz2015:capacitated_minimum_spanning_tree},
\citet{Silva2014:roots_non_linear_equations},
\citet{Stefanello2017:traffic_congestion_tolls},
\citet{Valente2009:single_machine_scheduling},
\citet{Londe2022:root_sequence_index}, and
\citet{Maecker2023:unrelated_parallel_machine_scheduling} or after, e.g., \citet{Goncalves2002:hybrid_assembly_line_balancing,Goncalves2005:hybrid_job_shop_scheduling,Goncalves2004:manufacturing_cell_formation}, \citet{Lopes2016:hub_location_routing}, \citet{Malve2007:max_min_lateness_parallel_batch_processing_dynamic_arrivals},
\citet{Ruiz2019:routing_capacity_distance},
and \citet{Valente2006:early_tardy_scheduling} the calculation of the fitness of the chromosome. 

%%%%%%%%% where in the BRKGA
%%%% inside decoder
%\cite{Valente2009:single_machine_scheduling} % before fitness calc
%\cite{Lopes2016:hub_location_routing} % sequentially after fitness
%\cite{Heilig2016:resource_management_cloud} % before fitness calc
%\cite{Stefanello2017:traffic_congestion_tolls} % before fitness calc
%\cite{Ruiz2019:routing_capacity_distance} % after fitness calc
%\cite{Goncalves2002:hybrid_assembly_line_balancing} % after fitness calc
%\cite{Goncalves2004:manufacturing_cell_formation} % after fitness calc
%\cite{Goncalves2005:hybrid_job_shop_scheduling} % after fitness calc
%\cite{Malve2007:max_min_lateness_parallel_batch_processing_dynamic_arrivals} % after fitness calc
%\cite{Fontes2007:concave_minimum_cost_network_flow} % before fitness calc
%\cite{Valente2006:early_tardy_scheduling} % after fitness calc
%\cite{Goncalves2011:foward_backward_improvement_constrained_scheduling} % before fitness calc
%\cite{Festa2013:data_clustering} % before fitness calc
%\cite{Moran-Mirabal2013:handover_minimization_networks} % before fitness calc
%\cite{Fontes2013:multi_pop_hop_constrained_trees_nonlinear} % before fitness calc
%\cite{Goncalves2014:akers_graphical_jobshop_scheduling} % before fitness calc
%\cite{Silva2014:roots_non_linear_equations} % before fitness calc
%\cite{Ruiz2015:capacitated_minimum_spanning_tree} % before fitness calc
%\cite{Londe2022:root_sequence_index} % before fitness calc
%\cite{Maecker2023:unrelated_parallel_machine_scheduling}	% before fitness calc

Outside of the decoding phase, the local search is usually applied to some of the best solutions on a specific number of generations. This is due to the fact that the application of local search on all individuals may be too computationally expensive for specific problems, especially if the mentioned local search has a large neighborhood. In this category, several papers apply local search on all offspring generated by the crossover process \citep{Buriol2005:weight_setting_problem_OSPF_routing,Buriol2007:OSPF_routing,Buriol2010:road_congestion,Ruiz2011:WSON_multilayer_network,Oliveira2017:hybrid_constrained_clustering,Silva2023:BRKGA_maximum_diversity,Brito2022:k_medoids_clustering}. Other papers only use local search after the evolutionary process is completed, often on the best incumbent solutions \citep{Roque2014:unit_commitment,Bernardino2018:traveling_purchaser,Biajoli2019:capacitated_facility_location,Mauri2021:multiproduct_facility_location,Chagas2021:bi_objective_traveling_thief}. Finally, some papers select special chromosomes to apply local search, usually only on selected generations \citep{Kashan2006:single_batch_processing,Lalla-Ruiz2016:quadratic_assignment,Londe2021:pnext_center,Carrabs2021:set_orienteering,Abreu2022:customer_order_scheduling,Japa2023:hybrid_brkga_hyperparameter_neural_nets}.

%%%% outside decoder
%\cite{Kashan2006:single_batch_processing} % best chrom of elite with set prob
%\cite{Chagas2021:bi_objective_traveling_thief} % if sol not dom other elites
%\cite{Londe2021:pnext_center} % best chrom of each gen
%\cite{Mauri2021:multiproduct_facility_location} % best sols at end
%\cite{Carrabs2021:set_orienteering} % all chrom all it, Mck on last it
%\cite{Lalla-Ruiz2016:quadratic_assignment} % each sol of elite only once
%\cite{Oliveira2017:hybrid_constrained_clustering} % all offspring
%\cite{Bernardino2018:traveling_purchaser} % all sol best pop and best sol find
%\cite{Biajoli2019:capacitated_facility_location} % 10 diff best sol at end
%\cite{Buriol2005:weight_setting_problem_OSPF_routing} % each sol fr crossover
%\cite{Buriol2007:OSPF_routing} % same as Buriol et al (2005)
%\cite{Buriol2010:road_congestion} % based on Buriol et al (2005)
%\cite{Ruiz2011:WSON_multilayer_network} % based on Buriol et al (2005)
%\cite{Roque2014:unit_commitment} % end of evo all elite
%\cite{Abreu2022:customer_order_scheduling}	% best sol elite set L interactions
%\cite{Brito2022:k_medoids_clustering}	% crossover, checks all possible combs
%\cite{Silva2023:BRKGA_maximum_diversity} % end of each gen to cur best sol not injected

One can also note that there is some variation among the explored neighborhoods. In fact, one can observe three options of moves among the used local searches. The first is the exchange between items of different categories, which is, by far, the most common approach. This includes the exchange of items in different categories, e.g.,  \citet{Biajoli2019:capacitated_facility_location,Buriol2010:road_congestion,Buriol2007:OSPF_routing,Buriol2005:weight_setting_problem_OSPF_routing};
\citet{Goncalves2002:hybrid_assembly_line_balancing};
\citet{Fontes2007:concave_minimum_cost_network_flow,Fontes2013:multi_pop_hop_constrained_trees_nonlinear};
\citet{Goncalves2004:manufacturing_cell_formation};
\citet{Goncalves2011:foward_backward_improvement_constrained_scheduling};
\citet{Carrabs2021:set_orienteering};
\citet{Kashan2006:single_batch_processing};
\citet{Londe2021:pnext_center};
\citet{Malve2007:max_min_lateness_parallel_batch_processing_dynamic_arrivals};
\citet{Mauri2021:multiproduct_facility_location};
\citet{Oliveira2017:hybrid_constrained_clustering};
\citet{Reis2011:OSPF_DEFT_routing_network_congestion};
\citet{Roque2017:multi_objective_unit_commitment};
\citet{Ruiz2015:capacitated_minimum_spanning_tree} and
\citet{Stefanello2017:traffic_congestion_tolls}, and creation or modification of the categories themselves, e.g.,\citet{Festa2013:data_clustering};
\citet{Goncalves2002:hybrid_assembly_line_balancing}, and
\citet{Ruiz2015:capacitated_minimum_spanning_tree}.

%%%%%%%%% ls neighborhood
%%%% exchange between items of different categories
%\cite{Kashan2006:single_batch_processing}
%\cite{Londe2021:pnext_center}
%\cite{Mauri2021:multiproduct_facility_location}
%\cite{Carrabs2021:set_orienteering}
%\cite{Stefanello2017:traffic_congestion_tolls}
%\cite{Oliveira2017:hybrid_constrained_clustering}
%\cite{Biajoli2019:capacitated_facility_location} %same as Mauri2021
%\cite{Ruiz2015:capacitated_minimum_spanning_tree}
%\cite{Goncalves2002:hybrid_assembly_line_balancing}
%\cite{Goncalves2004:manufacturing_cell_formation}
%\cite{Buriol2005:weight_setting_problem_OSPF_routing}
%\cite{Buriol2007:OSPF_routing}
%\cite{Fontes2007:concave_minimum_cost_network_flow}
%\cite{Valente2006:early_tardy_scheduling}
%\cite{Buriol2010:road_congestion}
%\citet{Reis2011:OSPF_DEFT_routing_network_congestion}
%\cite{Goncalves2011:foward_backward_improvement_constrained_scheduling} %not quite
%\cite{Festa2013:data_clustering}
%\cite{Moran-Mirabal2013:handover_minimization_networks}
%\cite{Fontes2013:multi_pop_hop_constrained_trees_nonlinear}
%\cite{Roque2014:unit_commitment}
%\cite{Ruiz2015:capacitated_minimum_spanning_tree}
%\cite{Malve2007:max_min_lateness_parallel_batch_processing_dynamic_arrivals}

The second possible approach is comprised of the swap of items in the same category. This move is more frequently used on problems whose solution is permutation-based, i.e., based on the ordering of items such as jobs in a schedule~\citep{Valente2009:single_machine_scheduling,Valente2006:early_tardy_scheduling,Abreu2022:customer_order_scheduling,Maecker2023:unrelated_parallel_machine_scheduling} or locations \citep{Bernardino2018:traveling_purchaser,Chagas2021:bi_objective_traveling_thief,Lalla-Ruiz2016:quadratic_assignment,Lopes2016:hub_location_routing}, often with adjacent swap moves.

%%%% interchange: swaps adjacent items
%\cite{Valente2006:early_tardy_scheduling}  % 2-opt % most effective
%\cite{Chagas2021:bi_objective_traveling_thief} % 2-opt on bi-objective
%\cite{Valente2006:early_tardy_scheduling} % 3-opt % not so effective
%\cite{Lopes2016:hub_location_routing} % 4-opt
%\cite{Lalla-Ruiz2016:quadratic_assignment} % 2-opt
%\cite{Bernardino2018:traveling_purchaser} % 2-opt
%\cite{Abreu2022:customer_order_scheduling}	% 2-opt 

The third commonly used neighborhood is composed of a greedy insertion or removal of the items of a solution. That is, the most advantageous item, in regards to the fitness value of the solution, is modified \citep{Bernardino2018:traveling_purchaser,Carrabs2021:set_orienteering,Goncalves2014:akers_graphical_jobshop_scheduling,Heilig2016:resource_management_cloud,Lopes2016:hub_location_routing,Silva2014:roots_non_linear_equations,Londe2022:root_sequence_index}.

%%%% largest insertion or removal: changes most costly item
%\cite{Valente2006:early_tardy_scheduling}  % not so effective
%\cite{Lopes2016:hub_location_routing}
%\cite{Carrabs2021:set_orienteering}
%\cite{Heilig2016:resource_management_cloud}
%\cite{Bernardino2018:traveling_purchaser}
%\cite{Goncalves2014:akers_graphical_jobshop_scheduling}
%\cite{Silva2014:roots_non_linear_equations}
%\cite{Londe2022:root_sequence_index} 

%%%%%%%%%%%%%%%% Tabu search

Several other search strategies used in BRKGA are derived from other well-known heuristics. The tabu search (TS) strategy first appeared in \citet{Glover1977:heuristics} and was first named as such in \citet{Glover1986:future_TS}. This method considers adaptive memory and responsive exploration, by forbidding certain moves in a neighborhood with either short- or long-term memory. The study of
\citet{Goncalves2014:akers_graphical_jobshop_scheduling} combines a tabu search with a local search based on Akers' graphical method~\citep{Akers1956:graphical_approach}.

%%%%%%%%%%%%%%%% VND-based searches

A variable neighborhood descent (VND) is a search strategy that explores different neighborhoods in a deterministic way \citep{Duarte2018:VND}. It is a variant of the variable neighborhood search (VNS) framework introduced in \citet{Hansen2005:VNS_VND}. As a search method, VND has been frequently hybridized with BRKGA. \citet{Pedrola2013:hybridizations_regenerator_placement_dimensioning} uses a VND after the application of a path relinking heuristic that explores, on some high-quality solutions, every $N$-move neighborhoods after a set number of generations. \citet{Stefanello2019:placement_virtual_machines} similarly applies the VND every generation, after the decoding step and the use of path relinking. The authors use four neighborhoods: shift, swap, chain2L, and chain3L, with the last two being chained shift moves.

\citet{Pessoa2018:flowshop_scheduling} applies a two-neighborhood VND inside the decoding process for a flowshop scheduling problem. Those neighborhoods are of the insert type, which moves jobs to new positions, and of the interchange type, which swaps two jobs. \citet{Andrade2019:flowshop_scheduling} uses a similar strategy but with a caveat: as the insert neighborhood acutely increases computational effort, it is restricted to only one interaction. The study of \citet{Soares2020:scheduling_identical_parallel} also inserts a VND into the decoder, with three search moves: insertion, interchange, and one-block grouping. The last neighborhood is specifically tailored to the problem characteristics.

Another hybridization of BRKGA and VND is shown by \citet{Zudio2018:BRKGA_VND_bin_packing} for a bin packing problem. The authors combine the VND with the crossover operator.
The combination of BRKGA and VND is also observed in \citep{Soares2022:resource_constrained_scheduling_setup} for a scheduling problem. This heuristic is applied to all individuals of the elite set, with four neighborhoods of insertion, exchange, relocation, and block grouping. In the same work, a simple local search is applied to all unexplored individuals of the non-elite set.

A random VNS (RVNS) is used by \citep{Schenekemberg2022:dial_a_ride}. After a set amount of generations, a label propagation method is used to identify communities of similar individuals. The RVNS is applied to the unexplored best solutions of each community, by randomly selecting one of the seven neighborhoods to explore. This approach is used with six neighborhoods by \citet{Chaves2024:adaptive_ftsp}. In this case, the application of the RNVS only occurs when the current generation is a multiple of $n/5$, with $n$ being the number of nodes in a TSP problem.
A VNS is used alongside the PAMPA framework proposed in \citep{Lu2023:additive_manufacturing_production}. To test the effectiveness of the BRKGA-based framework, it was applied to the solutions obtained by a VNS algorithm.

%%%%%%%%%%%%%%%% Clustering search

The clustering search approach \citep{Oliveira2013:clustering_search} identifies interesting individuals in the search space using clustering strategies and applies a local search to those individuals. The studies of \citet{Chaves2016:hybrid_tool_switches} and \citet{Silva2019:multicommodity_tsp} use a label propagation algorithm \citep{Raghavan2007:label_propagation} to find interesting solutions. This method constructs and iteratively labels a graph of the solution. The local search applied by \citet{Chaves2016:hybrid_tool_switches} considers two moves: the removal and insertion of items in the clusters, and the swap of items in different clusters. Meanwhile, \citet{Silva2019:multicommodity_tsp} uses an iterated local search (ILS) \citep{Lourencco2003:ils}, whose perturbation phase is based on swapping one-on-one items from different categories, and whose local search phase is a variable neighborhood descent \citep{Hansen2005:VNS_VND} with three neighborhoods: the two from \citet{Chaves2016:hybrid_tool_switches}, and a two-opt in which adjacent items are exchanged. In both papers, the use of the clustering search improves solution quality without significantly increasing computational times. 

%%%%%%%%%%%%%%%% adaptative large neighborhood search

An adaptive large neighborhood search (ALNS)~\citep{Pisinger2007:ALNS} is hybridized with BRKGA in \citet{He2019:scheduling_setup_times} for a scheduling problem with setup times. The ALNS is used after the decoding process, once for every solution that reaches a minimum fitness threshold. The ALNS uses five removal operators: random, minimal revenue, minimal unit revenue, maximal setup time, and worst sequence removal. The insertion operators are based on maximal revenue and maximal unit revenue.

%%%%%%%%%%%%%%%% Iterated greedy algorithm

An iterated greedy algorithm (IGA)~\citep{Ruiz2007:IGA_flowshop} is used in the intensification phase of the BRKGA proposed by \citet{Abreu2021:scheduling_routing_makespan}. This algorithm is applied after a set number of iterations to the best solution of that generation. It is based on the removal of a percentage of the operations of a solution, followed by the reconstruction of the solution. The hybrid approach had better performance than a pure BRKGA.

%%%%%%%%%%%%%%%% Iterative neighborhood search algorithm

In \citep{Xu2022:ABKRGA_local_search}, two Iterative Neighborhood Local Search Algorithms are introduced, one inexact and the other, exact. The inexact INLS is used on Individuals selected by a label propagation method after a set amount of generations. After the end of the evolutionary process, the exact INLS is used on the incumbent solution. The authors observe that this addition improves quality and performance in comparison with state-of-the-art methods.

%%%%%%%%%%%%%%%% Quick summary

As this section shows, there are several ways of combining BRKGA and local search heuristics. Generally, this hybridization manages to improve solution quality at an increase in computational effort. 

\subsection{Warm-start}

The classical BRKGA framework creates the initial population with randomly generated individuals. Nonetheless, the introduction of good solutions in the initial population has been noted to improve the performance of the algorithm. Studies in this category create one or several initial high-quality solutions and introduce them to the initially generated population as individuals.

Generally, this approach uses known heuristics from the literature to generate the initial solutions. \citet{Ericsson2002:Genetic_alg_OSPF} does this, using known heuristics to generate two solutions, while \citet{Buriol2005:weight_setting_problem_OSPF_routing} generates one initial solution using a simple heuristic. It is also done by \citet{Valente2006:early_tardy_scheduling}, but, in this case, three solutions are created. The authors show that the use of good individuals improves the performance of the algorithm. \citet{Pessoa2018:flowshop_scheduling} introduce in the initial population one solution from one effective heuristic from the literature. Meanwhile, \citet{Andrade2019:flowshop_scheduling} use diverse solutions of a known heuristic from the literature. A similar approach is used by \citet{Lopes2016:hub_location_routing}. \citet{Martarelli2020:feature_selection} creates half of the initial population randomly. In the other half, the first and second quartiles and the first solution from the third quartile receive solutions from three known heuristics. This strategy is shown to improve algorithm performance and convergence rate. The study of \citet{Abreu2021:scheduling_routing_makespan} uses a greedy insertion algorithm to create two solutions for the initial population. \citet{Chagas2021:bi_objective_traveling_thief} use known effective algorithms to generate solutions to both sub-problems (knapsack and TSP). Then, those solutions are combined to generate complete solutions for the traveling thief problem. In \citep{Xie2022:adaptive_brkga_cloud_workflow_scheduling}, three solutions come from algorithms from the literature, while all other initial individuals are obtained from a level heuristic. \citet{Londe2022:root_sequence_index} uses a customized heuristic to introduce one individual in the first population. \citet{Junior2023:a_brkga_laser_cutting} uses en eulerian heuristic to generate initial individuals for a TSP problem. Lastly, the insertion of solutions from two constructive heuristics in \citep{Silva2023:brkga_chordal_completion} is shown to increase algorithm performance.

An alternative to the use of heuristics to generate initial solutions is the use of mathematical models. \citet{Andrade2015:winner_auctions} introduces solutions from a linear relaxation of the MIP formulation of the problem. The model interactively fixes genes to create several solutions, and its use leads to better performance on all proposed decoders. \citet{Oliveira2018:car_rental} use three strategies to generate warm-start solutions: decomposing the complex integer non-linear programming model to a simpler one and solving it to optimality, relaxing the model's integrality constraints and solving it to optimality, and constructing naive solutions. The work of \citet{Chagas2020:traveling_salesman_loading} uses the model of a less constrained version of the problem to create a possible solution. This solution is the optimal solution of the relaxed model. The paper from \citet{Schuetz2022:robot_traject_planning} uses solutions from several alternative algorithms in their initial population. The methods include linear programming, quantum annealing, and greedy algorithms, among others.

An additional use of warm-start is when the proposed strategy demands specific characteristics improbable to occur in random individuals. The paper of \citet{Kashan2006:single_batch_processing} on single batch processing illustrates this situation, as the proposed BRKGA uses a decoding strategy that needs specific batch sizes to be effective, something that is improbable in random individuals. This means that, for that decoder, the entire initial population is generated by a robust heuristic. This heuristic aims to reduce the chance of BRKGA converging to bad solutions, by modifying the number of batches associated with them. \citet{Zhang2022:parallel_batch_scheduling_packing}, meanwhile, uses orthogonal design to create the initial population. This is shown to increase convergence rate and solution quality in comparison with random-generation of initial solutions.

One may observe that warm-start solutions have been shown to improve performance and convergence rate, independently of which methods are used to create the solutions. More than half of the studies with this hybridization use known algorithms from the literature, while the use of mathematical programming is also frequent.

\subsection{Other heuristics}
\label{Section:Results:Hybrids:Other}

Path Relinking (PR) \citep{Glover1997:PR} is an intensification strategy that explores the neighborhood in the path between two distinct solutions \citep{Ribeiro2012:PR}. \citet{Pedrola2013:hybridizations_regenerator_placement_dimensioning} uses this strategy alongside a VND. The evolutionary process of the BRKGA is stopped after a set number of generations when all solutions in the elite population have a chance of being selected for the so-called elite set. In this set, path-relinking and the VND can be applied to all solutions, with probability proportional to the distance between them and the guide solution. The PR used has a back-and-forth strategy, in which guide and base solutions change places after the first path is completed. \citet{Stefanello2019:placement_virtual_machines} also uses PR alongside VND. In this case, both are applied after the decoding step on all generations. If a better solution is found, then the corresponding chromosome is rewritten.
Both studies point out that the hybridization of path-relinking, VND, and BRKGA leads to better results without a severe increase in computational effort. Later, \citet{Schuetz2022:robot_traject_planning} uses PR to refine high-quality solutions of the last population of BRKGA.

The differential evolution (DE) algorithm is inspired by geometry instead of by nature \citep{Storn1997:differential_evolution}. \citet{Kong2020:parallel_scheduling_deterioration_maintenance} uses the JADE \citep{Zhang2009:jade} framework as the crossover operator of a BRKGA. This crossover is also of note as it considers three possible parents: one from the elite set, one from the non-elite set, and one randomly chosen among all individuals. The crossover probability comes from a normal distribution, while the mutation chance is derived from a Cauchy distribution. This addition causes longer running times but also improves solution quality and convergence rate in comparison with a pure BRKGA.

\citet{Queiroga2021:ils_scheduling} uses a dynamic programming (DP) approach \citep{Bellman1966:dynamic_programming} as the decoding process of a proposed BRKGA. Specifically, the authors use the approach proposed in \citet{Chou2008:DP}. This hybrid is shown to have worse performance than other approaches for the given problem.

The Flower Pollination algorithm \citep{Yang2012:FPA} is used in the crossover operator of \citet{Yu2023:supply_scheduling}. In it, there are three possible crossover combinations in each application. First, the global pollination strategy considers the combination of the current solution with the best known solution. Second, the local pollination scheme combines the current solution with one elite and one non-elite individuals. Third and last, a parameterized uniform crossover can be used. This approach had success in comparison with pure BRKGA and FPA.

\subsection{Multi-objective}

Multi-objective optimization problems bring challenges to the BRKGA framework, as the competition between objectives demands special adaptations to items such as decoding processes and chromosome ordering and selection. One such adaptation is the work of \citet{Zheng2015:multi_objective_container_loading}. In the chromosome ordering phase, their bi-objective algorithm uses a Pareto ranking process with the average of the weighted objectives. The weights used change at each generation, as they depend on their proportion among the current values of the complete population.

\citet{Tangpattanakul2015:multi_user_observation_scheduling,Tangpattanakul2015:scheduling_observations_satellite} also propose a bi-objective approach with modifications inside the chromosome ordering phase. The authors experiment with methods inspired by successful multi-objective algorithms, such as NSGA-II \citep{Deb2002:NSGAII}, SMS-EMOA \citep{Beume2007:SMS_EMOA}, and IBEA \citep{Zitzler2004:IBEA}, with results pointing to better computational times of the one inspired by IBEA. The authors also conclude that a hybrid decoder -- in which several algorithms are used as part of the decoding process, and the best solution among all is used as the chromosome fitness value -- is more effective than simpler decoding heuristics. Meanwhile, \citet{Roque2017:multi_objective_unit_commitment} indicates that a selection method based on NSGA-II can also be effective for a bi-objective BRKGA.

The bi-objective BRKGA of \citet{Cabo2018:bi_objective_scheduling} also uses the hybrid decoder strategy of \citet{Tangpattanakul2015:multi_user_observation_scheduling,Tangpattanakul2015:scheduling_observations_satellite}, but with a twist. In this case, one of the objective functions refers is the number of batches in the scheduling problem. After the decoding process is performed, this batch number is inserted into the last genes of the chromosome, while the other objective is used by the algorithm as the fitness of the chromosome.

Both papers of \citet{Huang2018:road_network_disruptions,Huang2020:bi_objective_road_disruptions} focus on the same bi-objective problem, but with different approaches. \citet{Huang2018:road_network_disruptions} uses an aggregated fitness function, with weights $\alpha$ and $1-\alpha$. This algorithm also updates the Pareto front every time a novel non-dominated solution is found. The other paper only has a modification regarding the aggregated fitness value of the chromosomes inside the decoding process. 

The studies by \citet{Raposo2020:network_reconfiguration_energy_loss},  \citet{Mikulski2021:energy_storage}, \citet{Bolsi2022:heuristic_workforce_allocation_scheduling}, and \citet{Marques2023:two_phase_multi_obj_UAV_routing} show the only BRKGAs adapted to more than two objective functions. \citet{Raposo2020:network_reconfiguration_energy_loss} consider a three-stage problem, that may be solved in two ways: the first stage is solved independently, then the two others are solved together, or all three stages are solved concurrently. The authors use the concept of Pareto dominance to select solutions and construct the Pareto front. The work by \citet{Mikulski2021:energy_storage}, meanwhile, modifies the elite set of the BRKGA for a multi-objective problem. In this case, the elite individuals are non-dominated solutions, while the non-elite chromosomes are dominated. This adaptation of the framework means that the selection and crossover operators are modified. The three objectives of \citet{Bolsi2022:heuristic_workforce_allocation_scheduling} are considered in a lexicographic manner, with three decreasing weights indicating preference. In \citep{Marques2023:two_phase_multi_obj_UAV_routing}, an aggregated fitness function is used for their three objectives. This means that a single-objective VNS could be used in their experiments with little modification.

\citet{Silva-Soto2021:timetabling_bus_lines} is the only study in which a bi-objective BRKGA approach is compared with a single-objective BRKGA. For the bi-objective algorithm, the authors combine BRKGA with a hierarchical non-dominated sorting algorithm for the chromosome sorting phase. For the comparison, the BRKGA uses an aggregated fitness function with weighted sums, in which $\alpha=0.5$. The hybrid approach was proven to be more efficient in finding the Pareto front than its competitor.

\citet{Chagas2021:bi_objective_traveling_thief} adapts the BRKGA to a bi-objective problem by hybridizing it with NSGA-II. This is done in the chromosome selection phase to join the strengths of both approaches for solving combinatorial optimization problems.

A novel multi--objective framework for BRKGA is introduced by \citet{Fontes2023:job_shop_transp_resources} and \citet{Fontes2023:bi_brkga_schedule_quay_cranes}. The multi--population BRKGA (mp--BRKGA) evolves $\Omega+\Pi$ populations independently and in parallel. Each $\Omega$ population considers only one of the $|\Omega|$ objectives, while the remaining $\Pi$ populations observe all objectives simultaneously. This algorithm borrows ideas from NSGA-II to create and rank Pareto fronts, but unlike NSGA-II, mp--BRKGA always allows offspring and mutant solutions to pass tho the next generations. The $\Omega$ populations use the classical BRKGA evolutionary process. Meanwhile, for each the $\pi \in \Pi$ populations the elite set is chosen from a pool of solutions, which consists of the best solutions of the previous $\Omega$ populations and the best individual of the given $\pi$ population. After a set number of generations, all $\Pi$ populations contribute with their best elite solutions to the others' pools, from which repeated chromosomes are removed. In tests, this approach was proven effective in comparison with NSGA-II and is able to cover the whole Pareto fronts of the instances successfully.

For this type of hybrid, some trends may be observed. Generally, BRKGA is adapted to bi-objective problems, with modifications to its chromosome ordering and selection phase and, more rarely, to the decoding process. The multi-objective algorithm is rarely compared with its single-objective counterpart and frequently uses concepts from effective multi-objective algorithms to improve performance. The use of aggregated fitness functions with weighted sums is also present in a third of the studies.

\subsection{Matheuristic}

There are several ways to combine mathematical formulations and heuristics to improve solution quality. A matheuristic is defined as the hybridization of metaheuristics and mathematical programming techniques \citep{Boschetti2009:matheuristics}. These methods use exact approaches as tools within the heuristic framework \citep{Fischetti2018:matheuristics}.
In BRKGA, mathematical programming is frequently applied to the decoding process. One such work was authored by \citet{Goncalves2011:lot_sizing_scheduling_backorders}, which uses a mathematical model as part of the decoder. After an individual is decoded, it goes to a surrogate model, which is then solved by an LP solver. The surrogate's objective value is used as chromosome fitness. A similar process is used by \citet{Goncalves2015:unequal_area_facility_layout}, with a slight difference. In this case, the LP model is only applied if the solution given by the chromosome is promising. The same idea of using a chromosome solution to obtain an easier surrogate model is also applied by \citet{Oliveira2018:car_rental}. This proposed algorithm applies a MIP solver to the best solution found by the BRKGA. \citet{Oliveira2019:matheuristic_car_rental_stochastic} uses a model inside the decoder to find the objective value of the second-stage problem, given the first-stage solution and possible scenarios. Lastly, \citet{Sun2022:bayesian_network_structure} uses the NOTEARS algorithm inside its decoder for Bayesina network structure learning. This algorithm considers the optimization of the dual version of the equivalent problem, which is also non-convex.

An alternative use of mathematical models inside BRKGA is in the form of exact local searches. An example of this is \citet{Correcher2017:berth_allocation}. The exact local search is applied to all individuals at the end of each generation for a berth allocation problem, as long as the time limit is not exceeded. This local search removes a cluster, then uses a MILP model to reallocate its items, while considering items in other clusters as fixed. This exact local search outperforms other methods and the matheuristic has better solution quality than a pure BRKGA. A similar strategy is used by \citet{Monch2018:matheuristic_batch_scheduling}. In this case, a model-based heuristic is applied if a solution is worse than a percentage of a reference value. Lastly, \citet{Pinto2021:maximum_quasi_clique_local_search} uses an exact local search. In this case, it is internal to the decoder, with a given probability of occurring in each decoding process. In comparison with a non-exact decoder, the matheuristic leads to consistently better solutions.

\citet{Andrade2017:feasible_mixed_integer} combines BRKGA with a feasibility pump heuristic to find feasible solutions to mixed integer programs. The proposed framework has three phases. BRKGA is used with a feasibility pump decoder in the evolutionary phase. After a few generations, the local MIP search phase occurs. It is applied to a pool of the best solutions, where solution quality is evaluated in regard to the number of non-integer variables. Meanwhile, the fixing phase reduces the problem's size by fixing variables that have the same values between relaxation and rounding. The use of the evolutionary phase is noted by the authors to lead to better results.

Lastly, \citet{Pinacho-Davidson2020:minimum_capacitated_dominationg_set} uses mathematical models in a novel way. In each generation, some of the solutions are selected to be merged. This selection may be elitist, with only elite solutions picked, or random. Afterwards, a MIP model is applied to the merged solution. The optimized solutions are then inserted back to the algorithm as a new chromosome, and the worst individuals in the population are eliminated. A solution-merging stateggy is also used in \citet{Lima2022:matheuristic_broadcast_time}. After some generations without improvement, solutions are merged and to create a sub--graph with the original vertices and the solutions' edges. An ILP model is used to optimize the problem induced by this sub--graph, and the obtained solution is added to a pool of best solutions.

Considering BRKGA-based matheuristics, there are some characteristics in common. One is using models inside the decoding process, which occurs in more than half of the studies. Moreover, comparisons between exact and non-exact heuristics are rare and only occur when the hybrid uses exact local searches.

\subsection{Machine learning}

Machine learning techniques may be used in conjunction with metaheuristics to improve performance by extracting useful knowledge within the search process \citep{Karimi2022:machine_learning}. One of its uses is for online parameter control, something that \citet{Chaves2018:capacitated_centered_clustering_local_search} applies to BRKGA. The proposed approach, called Adaptive BRKGA (A-BRKGA), updates parameters for population size, elite proportion, mutant proportion, crossover probability, and maximum number of generations at each iteration. It also uses two self-adaptive parameters, named $\alpha$ and $\beta$. The former indicates how many individuals in the elite restricted control list (RCL) evolve, while the latter is used as a probability of perturbing similar individuals. Parameters $\beta$ and crossover probability are included in the chromosome, thus evolving alongside the iterations. The crossover probability used is always the one belonging to the non-RCL parent.

Among the parameters, population size, mutant proportion, and $\alpha$ decrease with each iteration, while elite proportion increases. Maximum number of generations proportional to user-set parameter $\gamma$ and the minimum and maximum population sizes. In comparison to a pure BRKGA, A-BRKGA outperformed its competitor. 

With no changes, the same framework is used by \citet{Silva2019:multicommodity_tsp}. Again, the A-BRKGA performed better than a pure BRKGA, but the difference in results was not statistically significant. Later, it is also successfully applied in \citep{Junior2023:a_brkga_laser_cutting,Xu2022:ABKRGA_local_search}.

\citet{Schenekemberg2022:dial_a_ride} introduces BRKGA-QL, a novel online parameter control scheme. In it, a Q-Learning algorithm \citep{Watkins1992:q_learning} is used to control the same parameters as A--BRKGA with a Markov Decision Process in an environment in terms of actions, states, and rewards. The Q function is used to map state-action pairs to possible rewards, with an exponential decreasing $\eta$--greedy policy to chose following actions. This policy gives $1-\eta$ chance of choosing the action with highest reward to increase intensification, and $\eta$ probability of increasing diversification by choosing randomly.  In this case, $\eta$ gradually decays during the evolutionary process. This framework is also used in \citet{Chaves2024:adaptive_ftsp}, and in both cases was shown to be an interesting alternative in comparison with state-of-the-art methods.

Another approach using machine learning methods is  \citet{Pan2021:covid19_spread}. The proposed framework has a predictor creation phase, using a random forest algorithm. The results from this phase are then used with a BRKGA to formulate the ideal mitigation strategy for COVID-19 spread. 

\citet{Xie2022:adaptive_brkga_cloud_workflow_scheduling} proposes an adaptive decoding procedure. With three possible decoders, this procedure increases chance of a simpler decoding process being used by late evolution. This means that, in early generations, more complex decoding processes tend to be used and thus increase the chance of converging to local optima, while in late generations there is a higher chance of doing a global search in the solution space.

\subsection{Stochastic}

The first work in this category is \citet{Oliveira2019:matheuristic_car_rental_stochastic}. This work solves a two-stage stochastic optimization problem using two co-evolving populations. The first population evolves solutions for the first-stage problem. In the decoding process, the first-stage solutions proposed by the chromosome are fixed as part of an approximate LP model. This model is solved to optimality. The other population evolves scenarios, with a focus on diversification. The fitness function of the second population, thus, is a measure of how much the decoded scenario differs from the others. The link between populations is the calculation of total cost between first-stage solutions and scenarios.
The authors note that this approach does not consider probability distributions among the scenarios, but also that it provides the decision-maker with several solutions to a multi-stage problem with uncertainty. 

Later, the same authors introduce a method of scenario generation in \citep{Oliveira2022:diversity_scenario_gen}. In it, individuals represent possible scenarios and their fitness is measured by the variation of their impact on the first-stage solution, as calculated by crowding distance. There are three possible cases to create the scenarios: one for solution with perfect information, in which one first-stage solution is developed for each scenario; one for average scenario, in which one solution is generated and used to observe scenario impact on all generations; and one for moving average scenario, where a new first-stage solution is created for each new generation.

\subsection{Summary}

As shown in this section, the use of hybridization alongside BRKGA is a widespread practice that frequently improves both solution quality and convergence rate, but at the cost of higher computational effort. The explored papers also show that it is a versatile metaheuristic, capable of being combined in several ways with other frameworks. Of note is the use of local search heuristics as intensification strategies, done by a third of all studies and shown to improve performance.

There are some limitations observed in the studies presented in this section. The use of hybridization encumbers an already computationally heavy algorithm. This preoccupation with computational effort may be the reason for the limited number of papers in the machine learning and stochastic categories, as the techniques associated with both hybrids also have a high computational burden. Different functionalities inside the BRKGA framework might be a solution to mitigate this impact, as shown in the next section.

%------------------------------------------------------------%
%%%%%%%%%%%%%%%%%%%%%% Functionalities %%%%%%%%%%%%%%%%%%%%%%%

\section{New features added to BRKGA}
\label{Section:Features}

Several new features were added to the BRKGA framework in the ten years since its formal introduction. The difference between a hybridization and a feature is that the former defines items added from outside the BRKGA framework that demand modifications of the algorithm, while the latter is comprised of strategies adapted specifically for the framework. The most important features are described in this section.

\begin{description}
    \item[Island model.]
    The island model is a tool to prevent premature convergence of the algorithm \citep{Pandey2014:premature_convergence,Whitley1999:island_model}. It is based on a facet of Darwinian evolution, in which isolated populations evolve in different ways. Thus, this model involves the evolution of parallel populations, which exchange elite individuals after a set number of iterations. This procedure also improves individual variability \citep{Andrade2021:BRKGA_MP_IPR}.
    \item[Reset operator.]
    The application of a full population reset may be demanded if the algorithm cannot escape a local optimum for a high number of generations \citep{Toso2015:C++_app_BRKGA,Whitley1999:island_model}. This procedure re-initializes all chromosomes with randomly generated genes, destroying the benefits of convergence but potentially encountering new locations in the solution space. As this operator increases population diversity, it also prevents premature convergence \citep{Pandey2014:premature_convergence}.
    \item[Shake operator.] The shaking operator introduced in \citet{Andrade2019:flowshop_scheduling} aims to increase the diversity of the population while partially preserving some of the genes obtained in the evolutionary phase. This is done by partially re-initiating the population, applying random modifications to the elite individuals and re-initialising non-elite chromosomes with random keys. Thus, the structure of the elite chromosomes is partially preserved, and the diversity of the non-elite set is guaranteed.
    \item[Online parameter tuning.] The Adaptive BRKGA   \citep{Chaves2018:capacitated_centered_clustering_local_search} introduces online parameter tuning on the BRKGA framework. In this approach, the parameters relative to crossover probability, population size, elite and mutant proportions, and maximum number of generations are updated at each iteration. This is done to increase exploration in earlier generations while increasing exploitation in the later ones. The parameters are modified in one direction, i.e., the trajectory of the parameters is either increasing or decreasing over time. The same parameters are modified in the BRKGA--QL framework \citep{Schenekemberg2022:dial_a_ride}. In this approach, the Q-Leaning reinforced learning algorithm is used inside the BRKGA framework with possible increases and decreases. The chance of diversification decays exponentially during evolution in this approach.
    \item[Multi-parent crossover.]
    In the original BRKGA framework \citep{Goncalves2011:BRKGA}, the crossover operator uses two randomly chosen individuals as parents, one from the elite set and another from the non-elite set. The mating of the two parents is biased in favor of the genes belonging to the elite parent. \citet{Andrade2021:BRKGA_MP_IPR} introduces the multi-parent crossover. The algorithm uses $\pi_e$ elite parents and $\pi_t - \pi_e$ parents from the non-elite set, and the genes are selected with a fitness ranking-based bias function \citep{Bresina1996:stochastic_sampling}.
    \item[Implicit path-relinking.]
    The Implicit Path-Relinking procedure (IPR) is also introduced in \citet{Andrade2021:BRKGA_MP_IPR}. As an intensification strategy, the classical path-relinking explores the neighborhood obtained in the path between two distinct solutions \citep{Glover1997:PR,Ribeiro2012:PR}. The use of path-relinking is typically problem-dependent, but the implicit variant creates the procedure inside of the existing BRKGA solution space. The proposed method is applied alongside the island model so that base and guide solutions come from different populations. Those individuals may belong to either the elite set or be randomly selected among all chromosomes. \citet{Andrade2021:BRKGA_MP_IPR} introduces two types of this procedure, the permutation- and indicator-based IPR, whose application depends on the type of decoder. 
    \item[Multi-objective evolution.] \citet{Fontes2023:job_shop_transp_resources,Fontes2023:bi_brkga_schedule_quay_cranes} introduce the multi--population BRKGA (mp--BRKGA) for multi-objective combinatorial problems. In this approach, the $\Omega$ populations focuses on only one of the problem objectives, while the $\Pi$ populations concentrate on all existing objectives. The elite set of each of the $\pi \in \Pi$ populations are stored in a pool of solutions, comprised by the best current chromosome of $\pi$ and of all $\Omega$ populations. After a set amount of generations, solutions from all pools are mixed and evaluated.
    \item[Application programming interfaces.] Several papers introduce application programming interfaces for BRKGA. \citet{Toso2015:C++_app_BRKGA} presents a C++ application that handles most of the evolutionary processes. The user only needs to define the decoding function. This implementation may use the OpenMPI API to parallelize the decoding phase into multiple threads and can be downloaded at \url{http://github.com/rfrancotoso/brkgaAPI}. This API was used to develop a GNU-style dynamic shared Python/C++ library of BRKGA. This library was introduced in \citet{Silva2015:library_BRKGA}, where an extensive example of its utilization is detailed, alongside download and user instructions. \citet{Andrade2021:BRKGA_MP_IPR} presents the BRKGA-MP-IPR framework and corresponding API. This API, available in the C++, Python, and Julia languages, implements the multi-parent and IPR strategies discussed in this section, alongside the multi-population strategy.  \citet{Oliveira2021:scenario_generation} presents an API similar to the approach used in \citet{Oliveira2019:matheuristic_car_rental_stochastic}. This implementation co-evolves one population of scenarios and another of solutions for a two-stage stochastic problem. Finally, \citet{Oliveira2022:diversity_scenario_gen} introduces an API for scenario generation on two-stage problems, with three possible strategies for scenario evaluation.
\end{description}

One can note that the introduced features mainly focus on increasing population diversity, especially among elite solutions. Genetic algorithms tend to suffer from premature convergence, i.e., convergence to a sub-optimal solution \citep{Pandey2014:premature_convergence}. In BRKGA, this is related to a lack of diversity in the elite set \citep{Damm2016:field_tech_scheduling}.

%------------------------------------------------------------%
%%%%%%%%%%%%%%%%%%%%%% Issues/problems %%%%%%%%%%%%%%%%%%%%%%%

\section{Possible under-performing issues}
\label{Section:Underperforming}

One of the remarkable capabilities of BRKGA is its ability to extract and optimize hidden structures
in problems. This is possible due to the double elitist mechanism carried out during the evolutionary process. 
Notably, the standard BRKGA biased uniform crossover, when set correctly, can keep blocks of ``good'' 
genes (from the elite parents), possibly carrying optimal substructures from the problem. This mechanism is 
ideal for real-world problems that usually carry many hidden structures. However, vanilla BRKGA may lose 
performance on some classical problems, e.g., the simplest versions of set covering and routing 
problems, such as the Traveling Salesman Problem (TSP). 

This situation was previously noted in \citet{Resende2012:steiner_triple_covering} which applied BRKGA to solve the Steiner triple covering problem. 
Since the solution landscape 
is relatively flat for this problem, the authors reduced the crossover bias ($\rho$) and increased the mutant ratio ($p_m / p$)  
to inject more variability into the population. Though this led to long computational times, it enabled the BRKGA to explore a larger portion of the solution space than would be possible with a vanilla BRKGA using the usual parameter settings. We can observe a similar effect 
in~\cite{Andrade2015:winner_auctions}, where 
the winner determination problem in combinatorial auctions was modeled as a multidimensional knapsack 
problem, a relatively traditional problem with a flat response surface. While the BRKGA still produced 
better results than the other algorithms, the difference was not very large. This was also observed on the 2D and 3D bin packing problem \citep{Goncalves2013:2d_3d_bin_packing} where the authors modified the fitness function to include not only the number of bins but also the remaining available space in the bins thus unflattening the landscape. Lastly, we can note that the 
BRKGA does not perform well on the traditional Traveling Salesman Problem. Such a fact is due to the 
uniform crossover, which is inadequate for the problem. Indeed, several papers have developed 
special-purpose crossover methods such 
as~\cite{AlOmeer2019:Study_crossover_MTSP} 
and~\cite{Ahmed2020:Comparative_crossover_TSP} 
for the TSP, showing that such methods outperform standard ones.

Also, the double elitism comes with a price: premature convergence. This situation may benefit scenarios 
where we need to quickly respond to the user, such as real-time planning systems. However, premature convergence may lead to sub-optimal solutions when we can afford additional optimization time. Indeed, the 
 performance of BRKGA is greatly affected by its auxiliary perturbation procedures, such as reset and shaking. 
Although many papers do not emphasize the number of calls to methods, such as reset and shaking, this number of calls is frequent, and as Section~\ref{Section:Features} shows, novel features that affect population diversity are frequent in the literature. 
In~\cite{Andrade2019:flowshop_scheduling}, we 
can see the effect of the shaking procedure on the solution quality compared to a vanilla BRKGA. 

Therefore, properly tuning the parameters that control the population diversity as well as the thresholds to call these perturbation auxiliary methods is essential. 
The framework proposed here shows some important counters, such as the number of iterations between 
solution improvements, large offsets, and others that can be used to check the convergence.

\section{Possible directions for future research}
\label{Section:Agenda}

The development of BRKGA since its inception has encompassed several modifications and additions to the framework. It also sheds light on diverse strengths of the metaheuristic and details methods to increase performance and versatility. The literature survey also points out to several possible research avenues for BRKGA. This section details those worthwhile areas for future research efforts.

An area where those efforts can thrive is the development of features and generalizations of the framework for multi-objective problems and problems with uncertainty. The mp-BRKGA framework \citep{Fontes2023:job_shop_transp_resources,Fontes2023:bi_brkga_schedule_quay_cranes} is a potential starting point for multi-objective approaches, as it is shown to have success in the coverage of the whole Pareto frontier of the instances. However, while the mp-BRKGA framework was developed for multi-objective applications, both papers in which it is used study bi-objective problems. In fact, the use of BRKGA for problems with more than two objectives is scarce, and points out to a possible research avenue.
Meanwhile, for problems with uncertainty, solely two-stage problems were studied. In both cases, BRKGA was used to generate scenarios for the second-stage, either in parallel to the first-stage solution \citep{Oliveira2019:matheuristic_car_rental_stochastic} or using the first-stage solution to compute scenario quality \citep{Oliveira2022:diversity_scenario_gen}. Problems with uncertainty have several real-life applications, and the development of BRKGA-based frameworks to solve them would be noteworthy additions to the current literature.

Another possible research avenue is the application of hybridization, specially with matheuristics, machine learning (ML) methods, and other algorithms. As indicated in Section~\ref{Section:Hybrids}, presently there is no unified framework for the application of matheuristics in BRKGA. A possible reason is the fact that the most common use of mathematical models is inside the decoding process, which must be customized for each application. The use of exact constraints as part of the genetic operators (reproduction, crossover, mutation) has not been yet applied to BRKGA either.
Meanwhile, \citet{Karimi2022:machine_learning} defines eight possible uses of ML in metaheuristics: algorithm selection, fitness evaluation, initialization, operator selection, learnable evolution model, neighbor generation, parameter setting, and cooperation. The authors observe that those methods have already been used in, at least, one GA. However, currently only online parameter tuning \citep{Chaves2018:capacitated_centered_clustering_local_search,Schenekemberg2022:dial_a_ride} and operator selection \citep{Xie2022:adaptive_brkga_cloud_workflow_scheduling} were explored with BRKGA. Hybrids with machine learning methods are, thus, a promising area for study. 
Lastly, the hybridization of BRKGA and other single/multi--solution metaheuristic frameworks is shown to improve solution quality and convergence, e.g. the use of VND as local search or NSGA-II ranking methods in multi-objective approaches. This practice can be extended to other algorithms in the literature, such as GRASP, ILS, PSO, and others.

The exploitation of novel generalizations and hybrids with BRKGA implies in a demand for new APIs. The current APIs must be updated to reflect the new hybridization possibilities and state-of-the-art features, while others must be developed in popular programming languages. As Section~\ref{Section:Features} points out, the most recent complete API for BRKGA is the BRKGA-MP-IPR framework presented by \citet{Andrade2021:BRKGA_MP_IPR} in C++, Julia, and Python. However, only the C++ version is updated to reflect recent innovations. The Julia API has an older but functional version of the BRKGA-MP-IPR framework, while the Python version is incomplete. One possible future work is the further development of the Python API to observe the additions to the framework. Possibly a Python-based library could be developed for BRKGA. One can find implementations of BRKGA in Java, Rust, F\#, Matlab, and R, but those are not stable versions and/or do not consider present-day inclusions to the framework.
The definition of a committee to oversee API development for BRKGA is a possibility for the future of this metaheuristic.

The previous topics imply in novel applications for the framework. Section~\ref{Section:Applications} indicates that BRKGA versatility is a key feature to its performance, something that facilitates its use to diverse problems. One should note that novel applications can lead to new research avenues and novel hybridizations that, in turn, result in generalizations and new APIs. Thus, the development of BRKGA has the potential to turn into a virtuous cycle, where the algorithm evolves alongside the new research areas.

\section{Conclusion}
\label{Section:Conclusion}

This paper surveyed over~150 articles on the metaheuristic Biased Random-Key Genetic Algorithm (BRKGA). The first applications of this methodology were in the early 2000s \citep{Goncalves2002:hybrid_assembly_line_balancing,Ericsson2002:Genetic_alg_OSPF}, but the framework was only formally defined in 2011 \citep{Goncalves2011:BRKGA}. This metaheuristic has been extensively used in several optimization problems since its conception, proving to be reliable, efficient, and versatile. 

The first part of the survey discussed the~34 categories of problems that were studied with BRKGA since its first application. The prevalence of scheduling, network configuration, and location problems is of note, as those were some of the first studied with this metaheuristic \citep{Goncalves2002:hybrid_assembly_line_balancing,Ericsson2002:Genetic_alg_OSPF,Buriol2010:road_congestion}. Those are also problems in which BRKGA frequently performs the same or better with respect to solution quality than other approaches.
The second part of the survey indicates that hybrid BRKGA strategies are common, especially alongside local search heuristics. The use of hybridization is frequently concerned with increases in computational effort, as BRKGA is already a computational-heavy heuristic. Finally, the third part of the survey points out that the new features added to the framework usually focus on increasing population diversity to prevent premature convergence.

Possible avenues to broaden the scope of this study are the addition of other databases, such as Web of Science and arXiv, and the inclusion of thesis, dissertations, conference papers, and non-peer-reviewed material such as technical reports and preprints.

\section*{Acknowledgment}

Mariana A. Londe was supported by
the Brazilian
Coordination for the Improvement of Higher Level Personnel (CAPES)
under grant 88887.815411/2023. 
Luciana S. Pessoa was supported by
the Brazilian National Council for Scientific and Technological
Development (CNPq) under grant 312212/2021-6,
the Brazilian
Coordination for the Improvement of Higher Level Personnel (CAPES)
under grant 001, and
the Carlos Chagas
Filho Research Support Foundation of the State of Rio de Janeiro
(FAPERJ) under grants 211.086/2019, 211.389/2019, and
211.588/2021.

\clearpage
\appendix
\section{Supplementary material}

\setcounter{table}{0}
\setcounter{figure}{0}

\begin{figure}[htb]
    \centering
    \includegraphics[scale = 0.65]{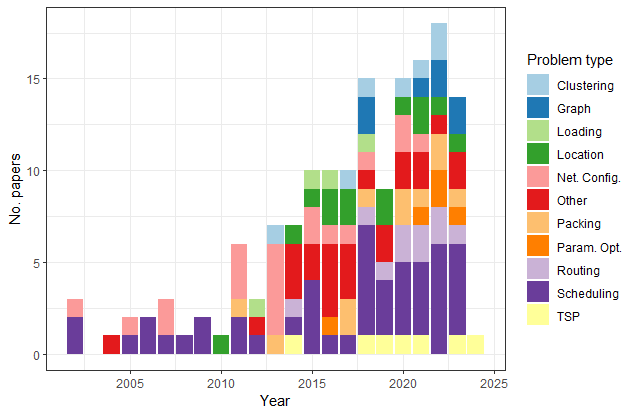}
    \caption{Number of papers per year and problem. Problem type ``other'' refers to problems studied in two or less articles.}
    \label{Figure:Results:Papers_problem_year}
\end{figure}

\begin{figure}[htb]
    \centering
    \includegraphics[scale = 0.65]{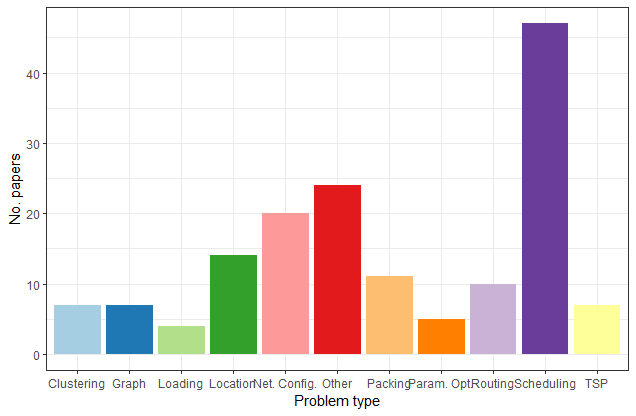}
    \caption{Number of papers per problem. Problem type ``other'' refers to problems studied in two or less articles.}
    \label{Figure:Results:Papers_problem_category}
\end{figure}

\begin{figure}[htb]
    \centering
    \includegraphics[scale = 0.65]{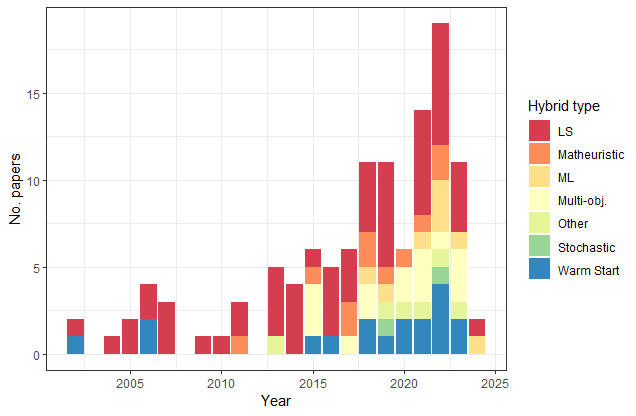}
    \caption{Number of papers per year and hybrid.}
    \label{Figure:Results:Papers_hybrids_year}
\end{figure}

\begin{figure}[htb]
    \centering
    \includegraphics[scale = 0.65]{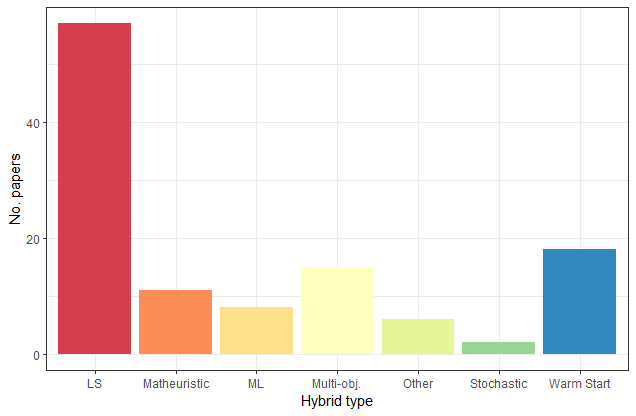}
    \caption{Number of papers per hybrid.}
    \label{Figure:Results:Papers_hybrids_category}
\end{figure}

% ---- Bibliography ----
%
\clearpage
\bibliographystyle{plainnat}
\bibliography{bibfiles/revBRKGA,bibfiles/bibliometric,bibfiles/new_papers}

\clearpage
\end{document}